\documentclass[11pt]{article}
\usepackage[margin=1in]{geometry}

\usepackage{microtype}
\usepackage{graphicx}
\usepackage{subcaption}
\usepackage{booktabs} %
\usepackage{siunitx}

\usepackage{xspace}
\newcommand*{\modelname}{\text{SRLM}\@\xspace}

\usepackage{wrapfig}

\usepackage[utf8]{inputenc} %
\usepackage[T1]{fontenc}    %
\usepackage{hyperref}       %
\usepackage{url}            %
\usepackage{booktabs}       %
\usepackage{amsfonts}       %
\usepackage{nicefrac}       %
\usepackage{microtype}      %
\usepackage[table]{xcolor}
\usepackage{enumitem}

\usepackage{subcaption}

\usepackage{newunicodechar}
\newunicodechar{≥}{\geq}

\definecolor{codegreen}{rgb}{0,0.6,0}
\definecolor{codegray}{rgb}{0.5,0.5,0.5}
\definecolor{codepurple}{rgb}{0.58,0,0.82}
\definecolor{backcolour}{rgb}{0.95,0.95,0.92}

\definecolor{custompink}{HTML}{DD99BB} 
\definecolor{customred}{HTML}{F66B2E} 
\definecolor{customyellow}{HTML}{F3AE36} 
\definecolor{customgreen}{HTML}{B9D5B1} 
\definecolor{customblue}{HTML}{4D9DE0} 
\definecolor{custompurple}{HTML}{AEADF0} 
\definecolor{customgray}{gray}{0.3}
\definecolor{custompurple}{HTML}{AEADF0} 
\definecolor{customwhite}{HTML}{fbf9f4}

\usepackage{newunicodechar}
\newunicodechar{≥}{\geq}

\definecolor{cyan10}{HTML}{E5F6FF}
\definecolor{cyan20}{HTML}{BAE6FF}
\definecolor{cyan60}{HTML}{0072c3}
\definecolor{cyan70}{HTML}{00539a}
\definecolor{cyan80}{HTML}{003a6d}
\definecolor{teal10}{HTML}{D9FBFB}
\definecolor{teal20}{HTML}{9EF0F0}
\definecolor{teal60}{HTML}{007d79}
\definecolor{orange10}{HTML}{FFF2E8}
\definecolor{orange20}{HTML}{FFD9BE}
\definecolor{orange60}{HTML}{ba4e00}
\definecolor{blue10}{HTML}{EDF5FF}
\definecolor{blue20}{HTML}{D0E2FF}
\definecolor{blue70}{HTML}{0043ce}
\definecolor{blue80}{HTML}{002d9c}
\definecolor{magenta10}{HTML}{FFF0F7}
\definecolor{magenta20}{HTML}{FFD6E8}
\definecolor{magenta30}{HTML}{ffafd2}
\definecolor{magenta50}{HTML}{ee5396}
\definecolor{magenta60}{HTML}{d02670}
\definecolor{magenta70}{HTML}{9f1853}
\definecolor{purple10}{HTML}{F6F2FF}
\definecolor{purple20}{HTML}{E8DAFF}
\definecolor{purple30}{HTML}{d4bbff}
\definecolor{purple70}{HTML}{8a3ffc}
\definecolor{rose10}{HTML}{FCF2ED}
\definecolor{rose20}{HTML}{F9D9D1}
\definecolor{rose60}{HTML}{ab5638}
\definecolor{rose70}{HTML}{853c27}
\definecolor{red10}{HTML}{FFF1F1}
\definecolor{red20}{HTML}{FFD7D9}
\definecolor{green10}{HTML}{DEFBE6}
\definecolor{green20}{HTML}{A7F0BA}
\definecolor{green70}{HTML}{0e6027}
\definecolor{green80}{HTML}{044317}
\definecolor{yellow10}{HTML}{fcf4d6}
\definecolor{yellow20}{HTML}{fddc69}
\definecolor{gray20}{HTML}{e0e0e0}
\definecolor{gray30}{HTML}{c6c6c6}
\definecolor{gray40}{HTML}{a8a8a8}
\definecolor{gray80}{HTML}{393939}

\definecolor{carbon-gray-10}{cmyk}{0.0, 0.0, 0.0, 0.04, 1.00}
\definecolor{carbon-gray-90}{cmyk}{0.0, 0.0, 0.0, 0.85, 1.00}

\usepackage{pifont}
\usepackage{amssymb}

\usepackage{wrapfig}

\usepackage{hyperref}

\usepackage{amsmath}
\usepackage{amssymb}
\usepackage{mathtools}
\usepackage{amsthm}

\usepackage[capitalize,noabbrev]{cleveref}

\theoremstyle{plain}

\theoremstyle{definition}

\theoremstyle{remark}

\usepackage{marginnote}

\usepackage[utf8]{inputenc} %
\usepackage[T1]{fontenc}    %
\usepackage{hyperref}       %
\usepackage{url}            %
\usepackage{booktabs}       %
\usepackage{amsfonts}       %
\usepackage{graphicx}       %
\usepackage{nicefrac}       %
\usepackage{microtype}      %
\usepackage{xcolor}         %
\usepackage{enumitem}

\usepackage{subcaption}
\usepackage{graphicx}

\usepackage[textsize=tiny]{todonotes}

\usepackage{newunicodechar}
\newunicodechar{≥}{\geq}

\definecolor{cyan10}{HTML}{E5F6FF}
\definecolor{cyan20}{HTML}{BAE6FF}
\definecolor{cyan60}{HTML}{0072c3}
\definecolor{cyan70}{HTML}{00539a}
\definecolor{cyan80}{HTML}{003a6d}
\definecolor{teal10}{HTML}{D9FBFB}
\definecolor{teal20}{HTML}{9EF0F0}
\definecolor{teal60}{HTML}{007d79}
\definecolor{orange10}{HTML}{FFF2E8}
\definecolor{orange20}{HTML}{FFD9BE}
\definecolor{orange60}{HTML}{ba4e00}
\definecolor{blue10}{HTML}{EDF5FF}
\definecolor{blue20}{HTML}{D0E2FF}
\definecolor{blue70}{HTML}{0043ce}
\definecolor{blue80}{HTML}{002d9c}
\definecolor{magenta10}{HTML}{FFF0F7}
\definecolor{magenta20}{HTML}{FFD6E8}
\definecolor{magenta30}{HTML}{ffafd2}
\definecolor{magenta50}{HTML}{ee5396}
\definecolor{magenta60}{HTML}{d02670}
\definecolor{magenta70}{HTML}{9f1853}
\definecolor{purple10}{HTML}{F6F2FF}
\definecolor{purple20}{HTML}{E8DAFF}
\definecolor{purple30}{HTML}{d4bbff}
\definecolor{purple70}{HTML}{8a3ffc}
\definecolor{rose10}{HTML}{FCF2ED}
\definecolor{rose20}{HTML}{F9D9D1}
\definecolor{rose60}{HTML}{ab5638}
\definecolor{rose70}{HTML}{853c27}
\definecolor{red10}{HTML}{FFF1F1}
\definecolor{red20}{HTML}{FFD7D9}
\definecolor{green10}{HTML}{DEFBE6}
\definecolor{green20}{HTML}{A7F0BA}
\definecolor{green70}{HTML}{0e6027}
\definecolor{green80}{HTML}{044317}
\definecolor{yellow10}{HTML}{fcf4d6}
\definecolor{yellow20}{HTML}{fddc69}
\definecolor{gray20}{HTML}{e0e0e0}
\definecolor{gray30}{HTML}{c6c6c6}
\definecolor{gray40}{HTML}{a8a8a8}
\definecolor{gray80}{HTML}{393939}

\definecolor{carbon-gray-10}{cmyk}{0.0, 0.0, 0.0, 0.04, 1.00}
\definecolor{carbon-gray-90}{cmyk}{0.0, 0.0, 0.0, 0.85, 1.00}

\usepackage{pifont}
\usepackage{amssymb}

\usepackage{wrapfig}

\usepackage{natbib}

\usepackage[most]{tcolorbox}
\usepackage{xcolor}
\usepackage{listings}
\newtcolorbox{promptbox}[2][]{
    colback=gray!8,
    colframe=gray!90,
    boxrule=0.8pt,
    arc=2pt,
    boxsep=3pt,
    left=8pt,right=8pt,top=8pt,bottom=8pt,
    fonttitle=\bfseries\small,
    title=#2,
    #1
}
\newtcolorbox{promptboxalt}[2][]{
    breakable,
    enhanced,
    colback=gray!8,
    colframe=gray!90,
    boxrule=0.8pt,
    arc=2pt,
    boxsep=3pt,
    left=8pt,right=8pt,top=8pt,bottom=8pt,
    fonttitle=\bfseries\normalfont,
    fontupper=\footnotesize,  
    title=#2,
    break at=-\baselineskip,      %
    height fixed for=none,        %
    pad at break*=3mm,           %
    overlay broken={\draw[gray!90,line width=0.8pt] 
        (frame.south west) -- (frame.south east);}, %
    #1
}

\usepackage{listings}
\usepackage{xcolor}
\lstdefinestyle{pythoncode}{
    language=Python,
basicstyle=\footnotesize\ttfamily,
    keywordstyle=\color{blue},
    commentstyle=\color{gray},
    stringstyle=\color{red},
    backgroundcolor=\color{gray!5},
    frame=single,
    frameround=tttt,
    breaklines=true,
    breakatwhitespace=true,
    showspaces=false,
    showstringspaces=false,
    tabsize=2,
    captionpos=b
}

\usepackage{algorithm}
\usepackage{algpseudocode}

\definecolor{darkgreen}{rgb}{0.0, 0.7, 0.0}

\title{Recursive Language Models Meet Uncertainty: The Surprising Effectiveness of Self-Reflective Program Search for Long Context}

\author{%
  Keivan Alizadeh\thanks{Equal contribution.} \hspace{0.5cm} 
  Parshin Shojaee$^{*}$
  \hspace{0.5cm}  
  Minsik Cho \hspace{0.5cm}
  Mehrdad Farajtabar \\
  [0.5cm] 
Apple
\date{}
}

\begin{document}
\setlength{\parindent}{0pt}
\maketitle

\begin{abstract}
\looseness=-1 
Long-context handling remains a core challenge for language models: even with extended context windows, models often fail to reliably extract, reason over, and use the information across long contexts.
Recent works like Recursive
Language Models (RLMs) have approached this
challenge by agentic way of decomposing long contexts into
recursive sub-queries through programmatic
interaction at inference. While promising, the success of RLMs critically depends on how these trajectories of context-interaction programs are selected, which has remained unexplored.
In this paper, we study this problem and introduce Self-Reflective Program Search for Long Context (\modelname), a framework that augments programming-based context interaction 
with uncertainty-aware self-reflection. \modelname leverages three intrinsic signals: self consistency, reasoning trace length, and verbalized confidence. These serve as complementary indicators of a model’s internal uncertainty, and the model uses them to evaluate and compare candidate context-interaction programs.
Extensive experiments across diverse benchmark datasets, context lengths, and backbone models, show that
SRLM consistently outperforms state-of-the-art
baselines, yielding up to 22\% improvement over
RLMs under the same time budget. 
Our findings show that recursion itself is not the primary driver of performance in RLMs, and a simple self-reflective program search can match or surpass RLM without requiring self-query or explicit recursion mechanisms. We find that for context lengths
within the model’s context window, RLMs with recursion
often degrade performance relative to the base
model, whereas SRLM yields consistent and robust gains across both short and long contexts.
We also find that RLM is less effective in tasks with semantically intensive nature, where heuristic program search is insufficient and broader contextual understanding is required, while self-reflection in \modelname provides a semantic
signal that better steers reasoning in these challenging long-context scenarios.
\footnotetext{Correspondence to \{pshojaee, kalizadehvahid, farajtabar\}@apple.com}
\end{abstract}

\begin{figure*}[t]
\centering
\includegraphics[width=\textwidth]{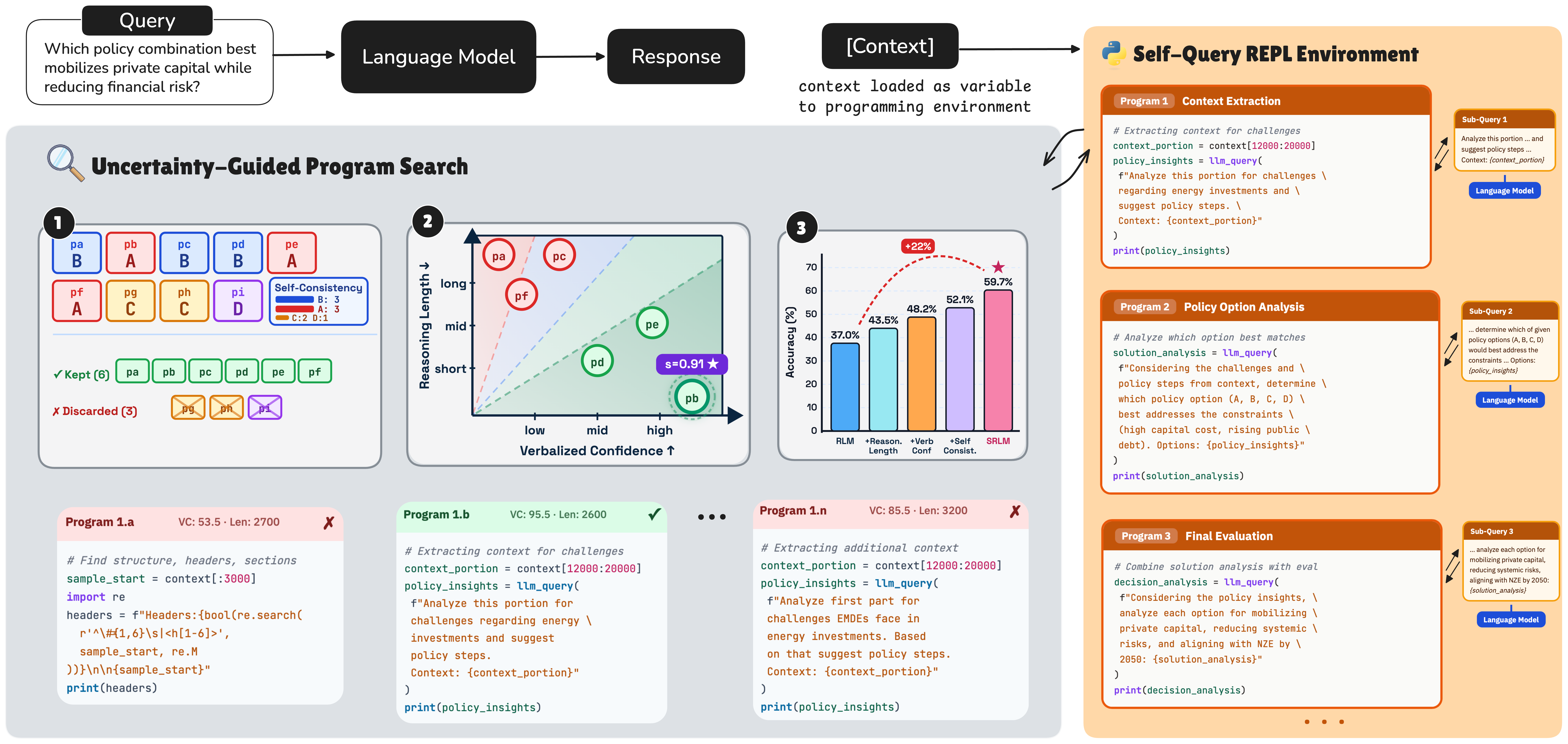}
\caption{ 
\looseness=-1 \textbf{Overview of \modelname}, 
a framework that augments programmatic context interaction reasoning with uncertainty-aware self-reflection. The language model operates in a self-query execution programming environment where the context is externalized as a variable, and generates programs that query and interact with context. Meanwhile, three complementary uncertainty signals (self-consistency, reasoning trace length, and verbalized confidence) are used to guide self-reflective programming trajectory selection without external supervision, enabling more robust and semantically grounded long-context reasoning.
}
\label{fig:main}
\end{figure*}

\section{Introduction}

Large language models are increasingly deployed in settings where long-context understanding is not optional but unavoidable. Modern applications from deep research agents \citep{huang2025deepresearchagentssystematic} and web browsing systems \citep{chen2025browsecomp} to coding assistants \citep{jimenez2024swebenchlanguagemodelsresolve} and self-improving agents \citep{zhang2025agenticcontextengineeringevolving} routinely demand reasoning over hundreds of thousands to millions of tokens spanning documents, logs, repositories, and interaction histories. 
Despite rapid progress in extending models' context windows, effective utilization of long contexts still remains challenging. Empirical studies show that even in frontier models with very large context windows, performance degrades with context length in ways that are well-documented but not yet solved: models lose track of salient details, fail to reliably extract, integrate, and reason over relevant information across distant positions, and are easily distracted by irrelevant content~\cite{liu2023lostmiddlelanguagemodels, hong2025context, du2025contextlengthhurtsllm}.

The research community has approached this challenge from several angles. 
One direction has been to target this problem at the model level for example through architecture sparsity mechanisms~\citep{tang2024quest,gao2024seerattention,lai2025flexprefill}, state-space models~\citep{gu2023mamba,dao2024transformers,waleffe2024empirical}, retrieval-based hybrid models~\citep{jin2024long,wang2023augmenting}, or KV cache compression~\citep{eyuboglu2025cartridges}, reducing the effective cost of processing long sequences.
Another direction has been at the data and training level where models are specifically trained on longer sequences or curating corpora that reward reasoning over long-horizons~\citep{fu2024data,zhao2024longskywork}.
A more recent promising direction treats long-context reasoning as a search problem at inference-time, leaving the model unchanged and instead restructuring how it interacts with context~\cite{wu2025resum,zhang2025agenticcontextengineeringevolving}. Chunking and summarization pipelines break long contexts into manageable pieces; retrieval systems surface relevant passages on demand; and agent-style frameworks issue iterative queries over the context, building up answers through a sequence of focused interactions.

Recursive Language Models (RLMs)~\citep{zhang2025recursive} represent the current state of the art in this inference-time context handling paradigm. 
Instead of processing long context with millions of tokens directly with model, RLM treats the context as an external variable within a programming environment, and allows the model to generate programs that query, slice, and recursively interact with the context. By externalizing context interaction with program execution, RLM has shown to extend the model’s effective reasoning horizon beyond what prompting typically allows. 

However, this framing introduces a largely unexplored dimension of the problem.
The quality of long-context reasoning in RLM is governed not only by the model’s capacity to process extended context, but also by the mechanism used to select trajectories of context-interaction programs. At each step, the model must decide which context segment to inspect, how to formulate intermediate self-queries, what sub-questions to pose, and how to aggregate these programming steps and partial results. The final prediction is therefore highly sensitive to the specific program trajectory instantiated during context interaction reasoning.
Despite this, RLM currently predominantly rely on fixed recursion schemes, lacking a principled mechanism for evaluating and selecting among alternative reasoning trajectories. 
This raises a central question:
\emph{Is recursion itself the key ingredient for long-context reasoning, or is the real bottleneck how we select among candidate interaction programs under uncertainty?}

In this work, we investigate this question and introduce Self-Reflective Program Search for Long Context Method (\textbf{\modelname}), a framework that augments programming-based context interaction with uncertainty-aware self-reflection~(\cref{fig:main}).
\modelname leverages three complementary signals (self-consistency, reasoning length, and verbalized confidence) as proxies for the model's internal uncertainty, 
enabling principled comparison of context-interaction trajectories through model's self-reflection without requiring external supervision.
Through extensive comparison experiments across diverse benchmarks, varying context lengths, and multiple backbone models, we observe that \modelname consistently outperforms state-of-the-art baselines, yielding up to 22\% improvement over RLM under the same wall-clock time budget.

Beyond empirical improvements, our analysis provides several insights into programming-based context-interaction frameworks like RLM and its key components. 
First, we find that recursion is not the primary driver of performance. A simple self-reflective program search can match or even surpass RLM without relying on explicit recursion or self-query mechanisms.
Second, the recursive self-query procedure is often more sensitive to context-length variations than self-reflection. In particular, when the context length falls within the model's native context window, recursive RLM reasoning can degrade performance relative to the base model, whereas \modelname yields more robust and consistent improvements across both short and long contexts.
Finally, we observe that RLM is less effective on semantically intensive tasks where heuristic program search is insufficient. In such settings, the uncertainty-aware self-reflection mechanism in \modelname provides a higher-level semantic signal that more effectively steers reasoning.
Together, these findings reposition recursion as one component of long-context reasoning rather than its defining feature, and suggest that uncertainty-aware self-reflection may serve as a simple yet effective alternative for building robust context-interaction frameworks.

More broadly, our goal in this paper is not just introducing a novel method but to better understand programming-based context-interaction frameworks and the role of their core components. Our study highlights the critical importance of programming trajectory selection in long-context interaction and suggests that improving how models explore and evaluate candidate interaction programs may be as important as extending context length itself.
We hope that these findings help guide the development of richer and more reliable long-context reasoning frameworks in future work.

Our key contributions are as follows:

\begin{itemize} [leftmargin=*,itemsep=0pt, topsep=0pt]
\looseness=-1 \item We introduce \modelname, a simple framework for long-context reasoning that augments programming-based context interaction with uncertainty-aware self-reflection. \modelname exploits three complementary uncertainty signals (self-consistency, reasoning trace length, and verbalized confidence) to enable principled comparison and selection of context-interaction programming trajectories.
\looseness=-1 \item We demonstrate that across diverse benchmarks, and multiple backbone models, \modelname consistently outperforms state-of-the-art baselines, achieving up to a 22\% improvement over RLM under the same wall-clock time budget.
\looseness=-1 \item We uncover that recursion is not the primary driver of RLM's performance, and a simple self-reflective program search can match or surpass recursion without the explicit self-query mechanism.
\looseness=-1 \item We find that RLM’s recursive procedure is sensitive to context length, mostly performing worse than the base model within the model's native context window, whereas \modelname delivers more robust improvements across both short and long contexts.
\looseness=-1 \item We identify a systematic failure mode of RLM on semantically intensive tasks and show that self-reflection provides a richer steering signal than heuristic-based recursive program search in these settings.
\end{itemize}

\section{Methodology}
\label{sec:method}

\subsection{Problem Formulation}
\label{sec:formulation}

\looseness=-1 Let $q$ denote a natural language query and $\mathcal{C} = (c_1, c_2, \ldots, c_N)$ a long context of $N$ tokens, where $N \gg L$ with $L$ being the model's effective context window. Rather than feeding $\mathcal{C}$ directly to model, we follow~\cite{zhang2025recursive} and treat context as an \emph{external variable} accessible within a sandboxed execution programming environment. A context-interaction program $p = (p_1, p_2, \ldots, p_T)$ is a sequence of $T$ executable operations, e.g., slicing, querying, or aggregating over $\mathcal{C}$, each generated autoregressively and executed in the REPL, producing an intermediate execution state: $e_t = \textsc{Exec}(p_t,\; e_{t-1},\; \mathcal{C}),$ where $e_0 = \varnothing$. The terminal step yields the program output $\text{out}(p) \in \mathcal{A}$ over answer space $\mathcal{A}$.
A key distinction from~\cite{zhang2025recursive} is that \modelname does \emph{not} require programs to instantiate explicit self-query sub-calls or recursive model invocations as tool calls. 
This decouples quality of context interaction from the structure of recursion, and shifts the focus of long-context reasoning improvement to the selection mechanism over candidate context-interaction program trajectories.

\subsection{\modelname: Self-Reflective Program Search for Long Context}

\looseness=-1
Given query $q$ and context $\mathcal{C}$, $K$ candidate programs are independently selected from the model policy $\pi_\theta$: $p^{(k)} \sim \pi_\theta(\cdot \mid q,\, \mathcal{C}), \quad k = 1, \ldots, K.$ 
Each $p^{(k)}$ constitutes a distinct reasoning trajectory over $\mathcal{C}$, differing in which context segments are inspected, how sub-problems are decomposed, and the confidence with which intermediate conclusions are drawn. 
We propose a self-reflective program search approach for long-context reasoning that draws on three complementary uncertainty signals: \emph{sampling-based uncertainty (self-consistency)}, \emph{semantic uncertainty (verbalized confidence)}, and \emph{behavioral uncertainty (reasoning trace length)}. Notably, all these three signals are derived from the model's own generation process, requiring no verifier, reward model, or external labeled data.

\subsubsection{Uncertainty Signals}
\looseness=-1 \paragraph{Sampling-based Uncertainty (Self-Consistency).}
As per~\cite{tao2025revisitinguncertaintyestimationcalibration}, a natural first-order uncertainty quantification arises directly from the sampling distribution over programs. Given $K$ independent draws from $\pi_\theta$, the empirical frequency of any candidate answer $a \in \mathcal{A}$ serves as an estimate of the model's marginal confidence in that answer, i.e., $\mathrm{prob}(a) = \frac{1}{K} \sum_{k=1}^{K} \mathbf{1}\bigl[\text{out}(p^{(k)}) = a\bigr] \;\approx\; \mathbb{P}_{\pi_\theta}\bigl[\text{out}(p) = a \mid q, \mathcal{C}\bigr]$.
The plurality answers $\hat{a} = \arg\max_{a \in \mathcal{A}}\, \mathrm{prob}(a)$ maximize this empirical confidence, and we retain the consistent candidate set as the subset of programs that agree with $\hat{a}$: $\mathcal{S} = \bigl\{p^{(k)} \in \mathcal{P} : \text{out}(p^{(k)}) = \hat{a}\bigr\} \subseteq \mathcal{P}$.
This step performs implicit verification through self-consistency~\citep{wang2022self}, however, self-consistency is a coarse uncertainty signal that operates only at the level of final outputs and is insensitive to the quality of the trajectory that produced them. Programs in $\mathcal{S}$ may share the same answer $\hat{a}$, yet may differ substantially in how they arrived at it: which context segments they inspected, how confidently they resolved each sub-problem, and how much deliberation they required. Selecting reliably among these candidates demands finer-grained uncertainty measures.

\looseness=-1 \paragraph{Semantic Uncertainty (Verbalized Confidence).}
Inspired by~\cite{xiong2023can}, to obtain a step-level semantic uncertainty signal, we elicit the model's own assessment of its confidence at each intermediate generation step $t$. Specifically, we append a structured instruction to the model's prompt, requiring it to report a confidence score for each step in a standardized format $\texttt{\{"confidence":}\; \nu_t^{(k)}\texttt{\}}, \quad \nu_t^{(k)} \in (0,\, 100],$ 
where the model is instructed to be precise and nuanced in its self-assessment. This elicitation yields a per-step confidence $\nu_t^{(k)}$ reflecting the model's self-assessed certainty over its intermediate conclusion at step $t$~\citep{xiong2023can}. Normalizing to the unit interval and aggregating in log-space over the full trace, we define the verbalized confidence score of program $p^{(k)}$ as $\mathrm{VC}(p^{(k)}) = \sum_{t=1}^{T^{(k)}} \log\!\left(\nu_t^{(k)}/100\right) \leq 0,$
where non-positivity follows from $\nu_t^{(k)}/100 \in (0,1]$, and values closer to zero indicate globally higher confidence across the trajectory. Unlike self-consistency, $\mathrm{VC}(p^{(k)})$ is a semantic uncertainty measure that captures how the model endorses each intermediate reasoning step as it progressively builds toward the final answer.
For more details of prompt used for this, check Appendix~\ref{app:confidence_prompt}.

\looseness=-1 \paragraph{Behavioral Uncertainty (Reasoning Length).}
While verbalized confidence relies on the model's explicit self-report at each step, we additionally exploit an implicit behavioral signal as the total token length of the generated trace. Let $\ell_t^{(k)}$ denote the number of reasoning and output tokens at step $t$; we define $\mathrm{Len}(p^{(k)}) = \sum_{t=1}^{T^{(k)}} \ell_t^{(k)}.$
We interpret this quantity as a proxy for epistemic effort. Intuitively, when a model is uncertain, it tends to generate longer, more deliberative traces, whereas confident and well-grounded reasoning is often associated with more concise outputs~\cite{devic2025trace, shojaee2025illusion}.
Importantly, trace length provides a signal complementary to verbalized confidence~\cite{devic2025trace}. Unlike self-reported confidence scores, it requires no explicit elicitation and is derived solely from observable generation statistics. As such, it offers an alternative fine-grained window into internal uncertainty that is not directly subject to miscalibration in the model’s stated confidence.

\subsubsection{Joint Uncertainty-guided Selection}
\looseness=-1 
The three uncertainty signals (self-consistency, verbalized confidence, and trace length) are complementary proxies of model uncertainty, each capturing a distinct aspect of the model's internal state. As our empirical results demonstrate (Section~\ref{sec:exp-ablation}), combining these signals yields a richer uncertainty characterization that more effectively guides program search over long-context interaction programs than any individual signal alone.
Within the consistent candidate set $\mathcal{S}$ (where self-consistency has already been enforced), we unify the remaining two signals into a joint uncertainty score of $s(p) = \mathrm{VC}(p) \cdot \mathrm{Len}(p)$ where lower values of $s(p)$ indicate better candidates.
By construction, $s(p) \leq 0$ since $\mathrm{VC}(p) \leq 0$ and $\mathrm{Len}(p) > 0$. Intuitively, this score penalizes programs that express low confidence or require excessively long reasoning traces---both indicators of uncertainty. The optimal program is then selected as $p^* = \arg\max_{p \in \mathcal{S}}\; s(p),$ with final prediction $\hat{y} = \text{out}(p^*)$.
Together, these three uncertainty signals form a coherent, self-reflective framework that effectively guides program search in \modelname without requiring any external supervision.

\section{Experiments}

\subsection{Datasets}
\looseness=-1 
Following~\cite{zhang2025recursive}, we evaluate \modelname on three benchmarks spanning diverse long-context reasoning tasks.
\textbf{BrowseComp+ (1K)}~\citep{chen2025browsecomp} is a multi-hop QA benchmark for DeepResearch~\cite{openai2025deepresearch} over a verified offline corpus of 1{,}000 documents, where each question requires piecing together evidence across multiple documents. Following~\cite{zhang2025recursive,sun2025scaling}, we evaluate on 150 randomly sampled instances and report accuracy.
\textbf{OOLONG (131K)}~\citep{bertsch2025oolong} requires transformation and aggregation of input chunks, scaling linearly in processing complexity with context length. We focus on the \texttt{trec\_coarse} split from OOLONG synthetic benchmark with context length $131$K (50 tasks),
and report scores following the original paper.
\textbf{LongBench-v2 CodeQA}~\citep{bai2024v2} is a multiple-choice code repository understanding benchmark requiring reasoning over long-context of files in a codebase (50 tasks).

\looseness=-1
Beyond this, we conduct extended evaluations targeting the core research questions of this study.
To characterize how context length affects \modelname and RLM, we evaluate on the \textbf{full OOLONG synthetic benchmark} (\texttt{trec\_coarse} split) across context lengths from $1$K to $4$M tokens (${\approx}650$ tasks, 50 per length).
To investigate the effect of task semantics and extend evaluation to tasks that by nature require more semantic understanding rather than heuristic search over context, we also evaluate on the \textbf{full LongBench-v2} benchmark across all domain categories beyond just CodeQA (${\approx}500$ tasks), including domains like single document QA, multi-document QA, long in-context learning, etc.
For more details on statistics, context length distributions, and category breakdowns of these datasets, check Appendix~\ref{app:datasets}.

\looseness=-1
\subsection{Baselines}
We compare against a comprehensive set of task-agnostic inference-time baselines following~\cite{zhang2025recursive}.
\textbf{Base LLM} processes the full context in prompt without any programmatic inference scaffolding.
\textbf{CodeAct (+BM25)}~\cite{wang2024executable} is a code-executing ReAct~\cite{yao2022react} agent that receives the full context directly and is additionally equipped with a BM25 retriever~\cite{robertson2009probabilistic} for context search as per~\cite{zhang2025recursive, jimenez2023swe, chen2025browsecomp}.
\textbf{CodeAct (+sub-calls)} ablates the effect of context offloading as a variable in REPL by augmenting the CodeAct baseline with the ability to invoke sub-calls from the language model.
\textbf{Summary agent} also follows~\cite{sun2025scaling,wu2025resum,yu2025memagent} and iteratively compacts and summarizes context as the model window fills, chunking documents that exceed the context limit.
\textbf{RLM}~\citep{zhang2025recursive} is the current state-of-the-art approach, externalizing context as a variable in a REPL environment and issuing recursive self-queries; we consider both the recursive variant (depth one) and the \textbf{no sub-calls} variant that disables this self-query procedure.
For each comparison across baseline methods, we use the same backbone models and sampling parameters.

\looseness=-1
\subsection{Experimental Setup}
In our experiments, we use two backbone LLMs: the open-weight Qwen3-Coder-480B-A35B~\cite{qwen3technicalreport} and GPT-5~\cite{singh2025openaigpt5card} with medium reasoning effort, with GPT-5-mini as the sub-model for the recursive calls (as per~\cite{zhang2025recursive}).
\modelname operates in the same REPL environment as RLM and uses $K{=}8$ candidate trajectories for uncertainty-guided program search, with uncertainty signals defined in Section~\ref{sec:method}.
To ensure fair wall-clock time comparison across methods, we impose execution time limits of 600 seconds per each step of trajectory for all runs.
We set a maximum of 30 program interaction steps and a maximum generation length of 260K tokens for Qwen3-Coder-480B, with default API parameters for GPT-5 and GPT-5-mini calls.
For verbalized confidence elicitation, we augment the original RLM prompt (As in~\cite{zhang2025recursive}) with a suffix requesting the self-report of internal confidence in a structured format, without modifying any other part of the prompt or reasoning procedure (see Appendix~\ref{app:confidence_prompt} for details).
For final answer evaluation, we also use \text{GPT-5-mini as a judge} across all datasets to robustly assess the correctness (check Appendix~\ref{app:judge_prompt} for details).

\begin{table*}[t]
\centering
\caption{
\looseness=-1
Performance comparison of \modelname{} against baselines on long-context benchmarks from~\cite{zhang2025recursive}. Results report accuracy (\%) on LongBench-v2 CodeQA, BrowseComp+ (1K documents), and OOLONG (131K tokens). \modelname{} consistently outperforms all baselines, achieving up to 22\% improvement over RLM. $^*$ indicates context overflow; $^\dagger$ indicates our replication of results; and \textbf{bold} shows best result per LLM backbone. 
}
\resizebox{\textwidth}{!}{
\begin{tabular}{lccc}
\toprule
\textbf{Model} & \textbf{LongBench-v2 (CodeQA)} & \textbf{BrowseComp+(1K)} & \textbf{OOLONG (131K)} \\
\hline
{Task Length } $N$ (tokens) & 23K-4.2M & 6M-11M & 131K \\
\midrule
\multicolumn{4}{c}{\emph{Qwen3-Coder-480B}} \\
\hline
Base Model & $20.0^*$ & $0.0^*$ & $36.0$ \\
CodeAct (+ BM25) & $24.0^*$ & $12.7$ & $38.0$ \\
CodeAct (+ sub-calls) & $26.0^*$ & $0.0$ & $32.0$ \\
Summary agent & $50.0$ & $38.0$ & $44.1$ \\
\rowcolor{cyan!15}RLM & $59.8^\dagger$ & $37.1 ^\dagger$ & $45.7^\dagger$ \\
\rowcolor{cyan!10}RLM (no sub-calls) & $53.8^\dagger$ & $36.3^\dagger$ & $39.1^\dagger$ \\
\hline
\rowcolor{magenta!20}\modelname & \textbf{64.9} {\footnotesize(\textcolor{darkgreen!100}{$\uparrow$5.1})}& \textbf{59.7} {\footnotesize(\textcolor{darkgreen!100}{$\uparrow$22.6})}& \textbf{51.8} {\footnotesize(\textcolor{darkgreen!100}{$\uparrow$6.1})}\\
\rowcolor{magenta!10}\modelname (no sub-calls) & 59.0 {\footnotesize(\textcolor{darkgreen!100}{$\uparrow$5.2})}& 50.1 {\footnotesize(\textcolor{darkgreen!100}{$\uparrow$13.8})} & 45.9 {\footnotesize(\textcolor{darkgreen!100}{$\uparrow$6.8})} \\
\midrule
\multicolumn{4}{c}{\emph{GPT-5}}\\ 
\hline
Base Model & $24.0^*$ & $0.0^*$ & $44.0$ \\
CodeAct (+ BM25) & $22.0^*$ & $51.0$ & $38.0$ \\
CodeAct (+ sub-calls) & $24.0^*$ & $0.0^*$ & $40.0$ \\
Summary agent & $58.0$ & $70.5$ & $46.0$ \\
\rowcolor{cyan!15}RLM & $59.5^\dagger$  & $86.0^\dagger$ & $53.0^\dagger$ \\
\rowcolor{cyan!10}RLM (no sub-calls) & $65.2^\dagger$ & $89.7^\dagger$ & $50.5^\dagger$ \\
\hline
\rowcolor{magenta!20}\modelname & 68.9 {\footnotesize(\textcolor{darkgreen!100}{$\uparrow$9.4})}& 92.4 {\footnotesize(\textcolor{darkgreen!100}{$\uparrow$6.4})}& \textbf{65.5} {\footnotesize(\textcolor{darkgreen!100}{$\uparrow$12.5})} \\
\rowcolor{magenta!10}\modelname (no sub-calls) & \textbf{74.1} {\footnotesize(\textcolor{darkgreen!100}{$\uparrow$8.9})}& \textbf{94.6} {\footnotesize(\textcolor{darkgreen!100}{$\uparrow$4.9})}& 60.7 {\footnotesize(\textcolor{darkgreen!100}{$\uparrow$10.2})} \\
\bottomrule
\end{tabular}
}
\label{tab:mainres}
\end{table*}

\subsection{Main Results}
\label{sec:exp-main}
\looseness=-1 
Table~\ref{tab:mainres} compares \modelname with RLM and other baselines on the long-context benchmarks from~\cite{zhang2025recursive}. Across all datasets and both backbone LLMs, \modelname consistently performs the best, improving over the previous state-of-the-art RLM by up to 22\%. 
Looking more closely, we find that the effect of recursion is inconsistent across backbone models. 
For example, with Qwen3-Coder-480B, recursion helps both RLM and \modelname, suggesting that decomposing the context into smaller sub-problems through model's self-query procedure as tool call can support long-context handling in Qwen backbone. However, under GPT-5, recursion hurts performance and the variants without sub-calls outperform their recursive counterparts in most cases. This might indicate that when the backbone model is already strong at long-context reasoning, explicitly using self-query tool calls for context-interaction may be unnecessary or even disruptive.
In contrast, the self-reflection mechanism in \modelname provides stable improvements across both of these backbones. Even without any sub-calls, \modelname often outperforms recursive RLM.
This suggests that guiding through model internals and self-reflection may matter more than guiding through explicit recursive tool calls for the scope of long context. We analyze these behaviors in greater detail in Section~\ref{sec:exp-context}–\ref{sec:exp-semantic}.

\subsection{Robustness Across Context Lengths}
\label{sec:exp-context}
\begin{figure*}[t]
\centering
\begin{minipage}{\textwidth}
\centering
\begin{subfigure}[t]{0.49\linewidth}
\centering
\includegraphics[width=\linewidth]{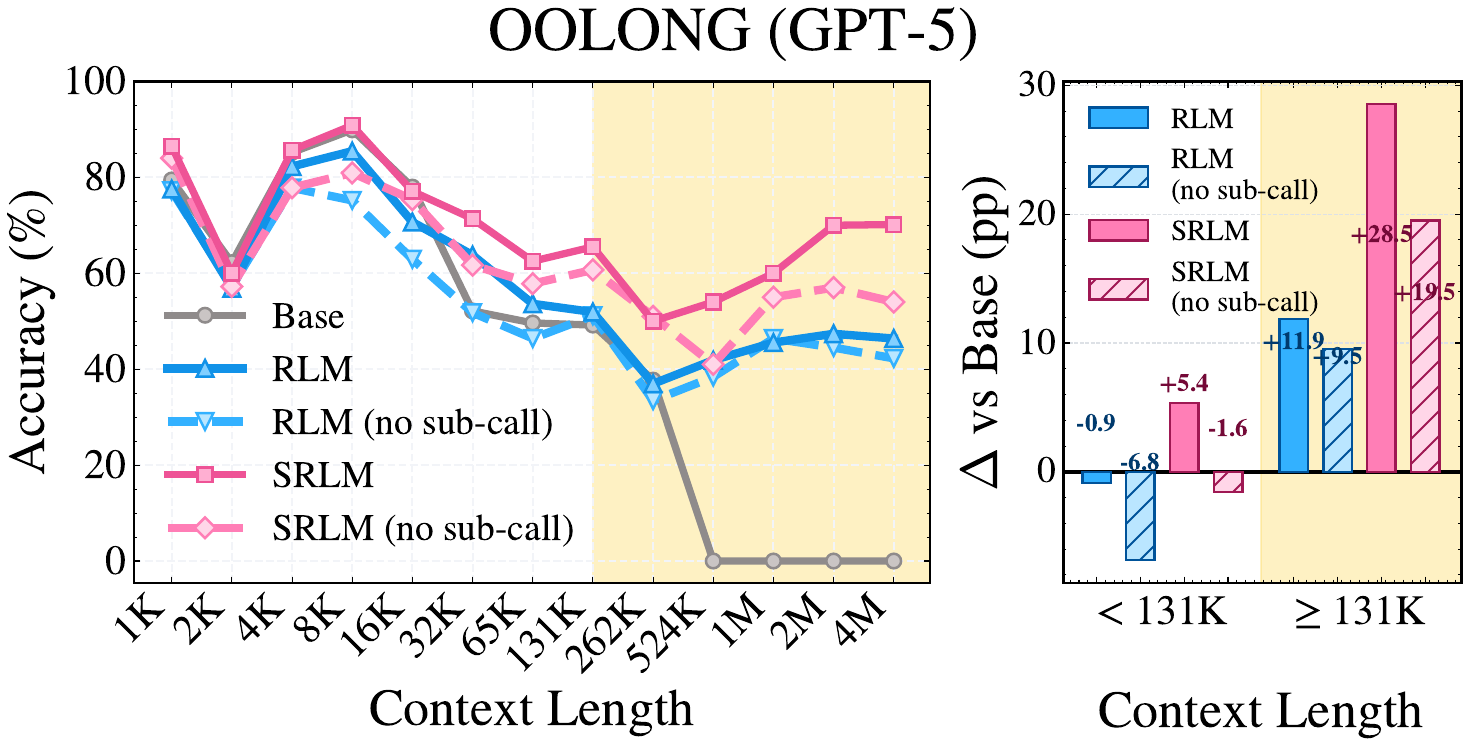}
\end{subfigure}
\hfill
\begin{subfigure}[t]{0.49\linewidth}
\centering
\includegraphics[width=\linewidth]{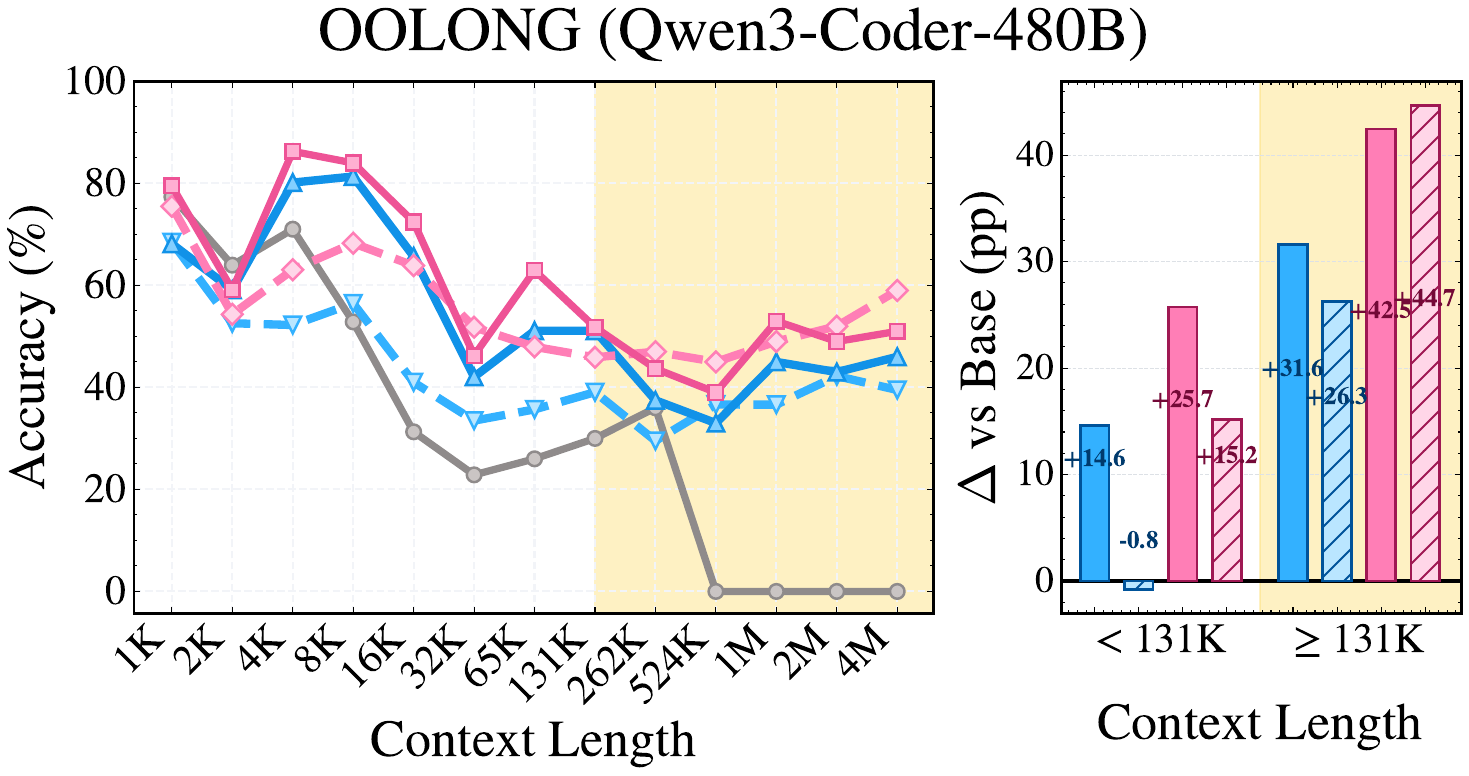}
\end{subfigure}

\vspace{0.5em}

\begin{subfigure}[t]{0.49\linewidth}
\centering
\includegraphics[width=\linewidth]{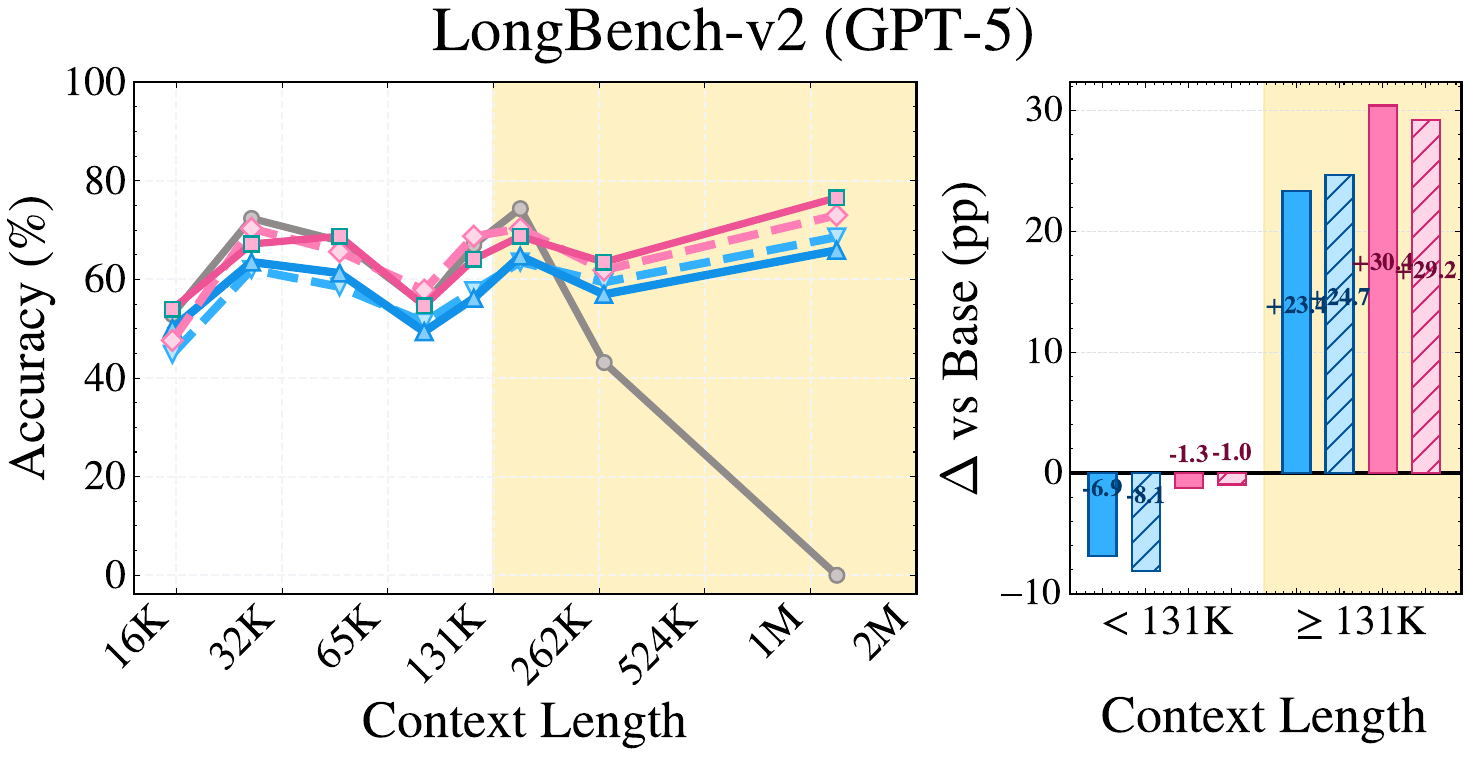}
\end{subfigure}
\hfill
\begin{subfigure}[t]{0.49\linewidth}
\centering
\includegraphics[width=\linewidth]{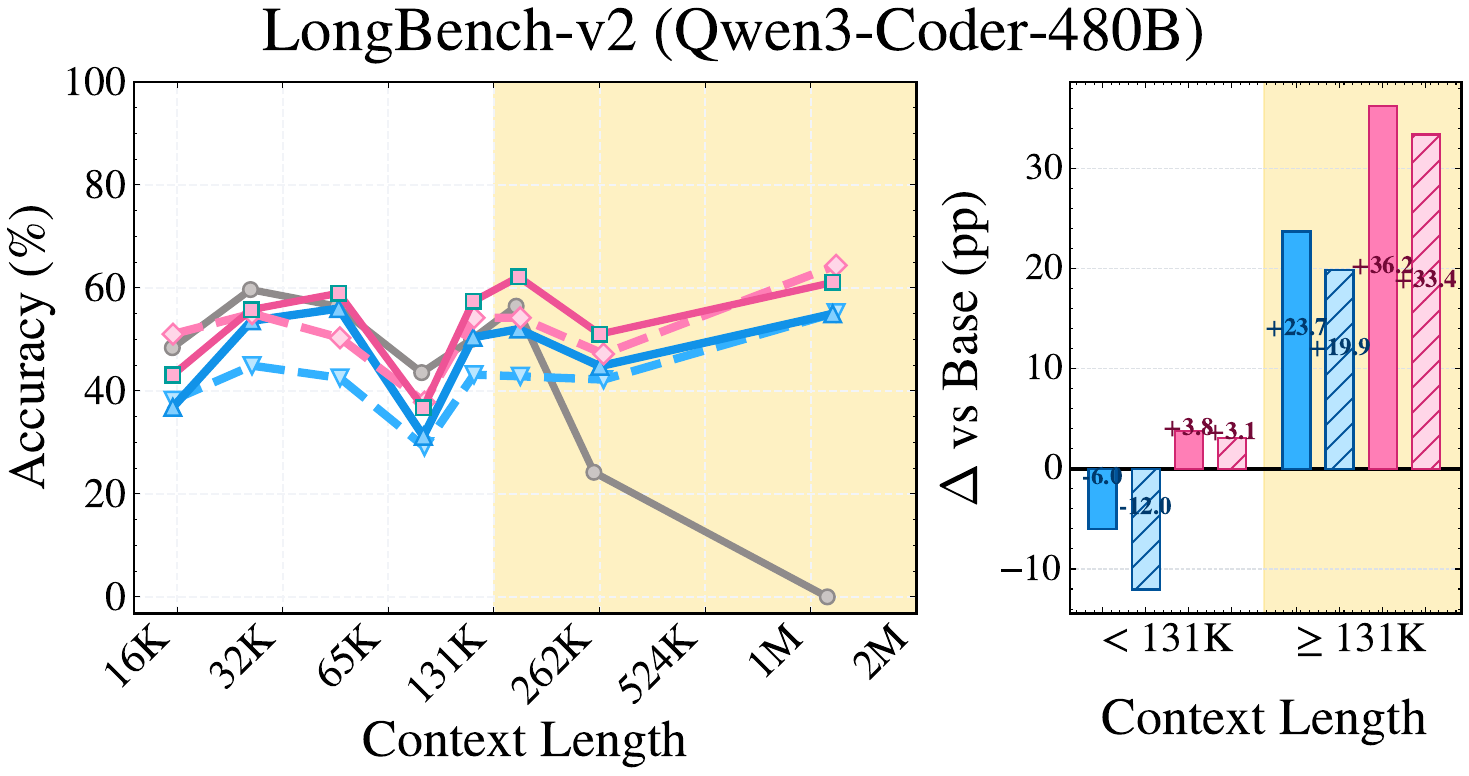}
\end{subfigure}
\end{minipage}
\caption{
Performance across context lengths on OOLONG and LongBench-v2 Full datasets: 
Line plots show accuracy of \modelname, RLM, and the base LLM across context from thousands to millions of tokens using GPT-5 \textbf{(left)} and Qwen3-Coder-480B \textbf{(right)} backbones. Bar plots show the average performance gain over the base model, separated into contexts within (${<}131$K) and near/beyond (${\geq}131$K) the native context window.
}
\label{fig:context}
\end{figure*}

Next, we investigate how context length affects the behavior of each method. To this end, we run additional experiments on the full LongBench-v2 and OOLONG datasets, covering contexts from thousand to millions of tokens. 
Figure~\ref{fig:context} compares \modelname, RLM, and the base LLM across context lengths. We also demonstrate the performance gap relative to the base model ($\Delta$ vs.\ base), separating results for shorter contexts well within the model's context window (${<}131$K), and longer contexts near or beyond the context limit (${\geq}131$K).
From these results, we observe several interesting patterns.
First, the advantage of \modelname becomes more pronounced as context length increases. On longer contexts, \modelname consistently provides better gains over the base model than RLM.
Second, RLM is noticeably more sensitive to context length. On shorter contexts (for example <$131$K), RLM often underperforms the base model, indicating that recursive decomposition may not be effective on all contexts and can introduce unnecessary overhead when the context is already manageable. In contrast, \modelname remains robust and provides more consistent gains over the base model on both short and long contexts. For more detailed results over backbones and tasks, check Appendix~\ref{app:context_details}.

\subsection{Is Recursion the Primary Driver of Performance in RLMs?}

One of the key motivations and a central question to this study is whether recursion is the main source of gains in recursive language models (RLMs), particularly in long-context settings. Understanding this can help toward guiding the design of more effective frameworks.
Conceptually, recursion can be viewed as a form of inference-time scaling through model as a tool use, i.e., the model decomposes the problem into sub-queries and recursively calls itself as a tool to interact with different parts of the context. 
In contrast, our self-reflective variant (\modelname without sub-calls) performs inference-time scaling through model's internals. Instead of explicitly issuing recursive tool sub-calls, it relies on implicit uncertainty-guided self-reflection to revise and refine its context-interaction programs.
To study the above research question, we focus on the comparison of context-interaction recursive programming (RLM with sub-calls) and self-reflective programming (\modelname without sub-calls) on long-context settings (${ \geq }131$K tokens) across various datasets and backbones.

\begin{wrapfigure}[17]{r}{0.45\linewidth}
\centering
\vspace{-1.0em}
\includegraphics[width=\linewidth]{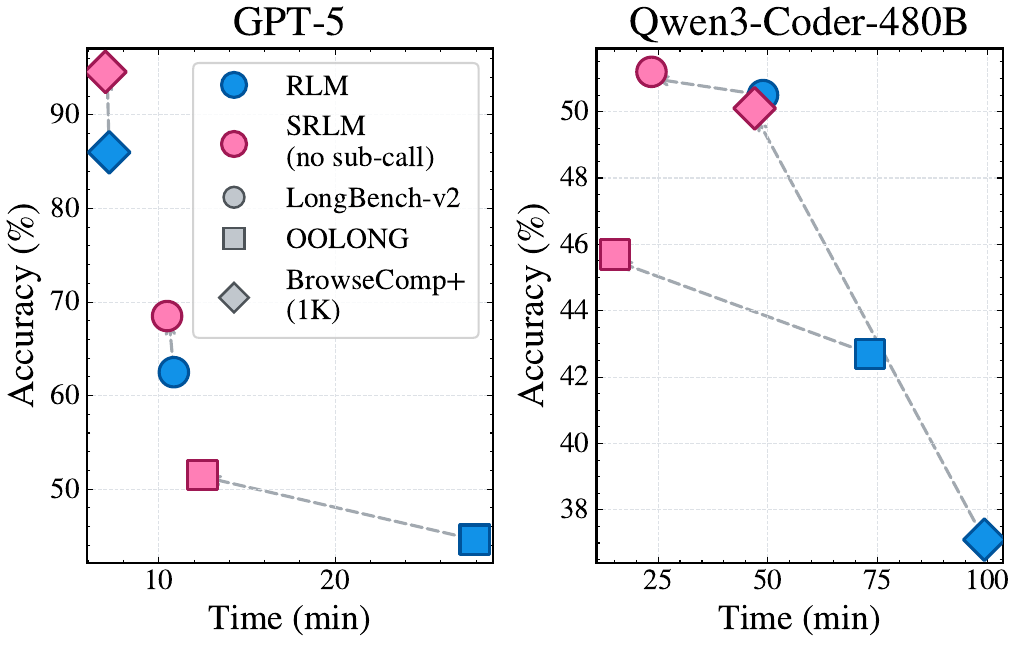}
\caption{
Accuracy versus cost pareto comparison of RLM and \modelname (no sub-call) on long-context settings of benchmarks 
under GPT-5 (\textbf{left}) and Qwen3-Coder-480B (\textbf{right}). 
}
\label{fig:pareto}
\end{wrapfigure}
As shown in Figure~\ref{fig:pareto}, self-reflection can actually outperform recursion in both performance and cost (wall-clock time)  under long-context settings. Notice that since all the trajectories are executed in parallel, the wall-clock time of \modelname doesn't significantly increase over RLM which runs only one trajectory. This suggests that recursion may not be the best strategy for inference-time scaling in long-context interactions, as explicit self-querying and sub-calls introduce additional overhead with only marginal performance gains.  
This trend is also reflected in Table~\ref{tab:mainres} results. In cases where recursion helps, most of the gains over the base model come from the programmatic context-interaction procedure rather than recursion. For example, on LongBench CodeQA with Qwen3-Coder-480B, performance improves from $20$ to $53.8$ with RLM without sub-calls, and only further to $59.8$ with recursive sub-calls which can be obtained with \modelname without sub-calls as an alternative inference time scaling method, indicating that recursion contributes only marginal gains in long-context frameworks.

\subsection{Task Semantics and Limits of Recursion}
\label{sec:exp-semantic}

\begin{wrapfigure}[19]{r}{0.45\linewidth}
\centering
\vspace{-1.0em}
\includegraphics[width=\linewidth]{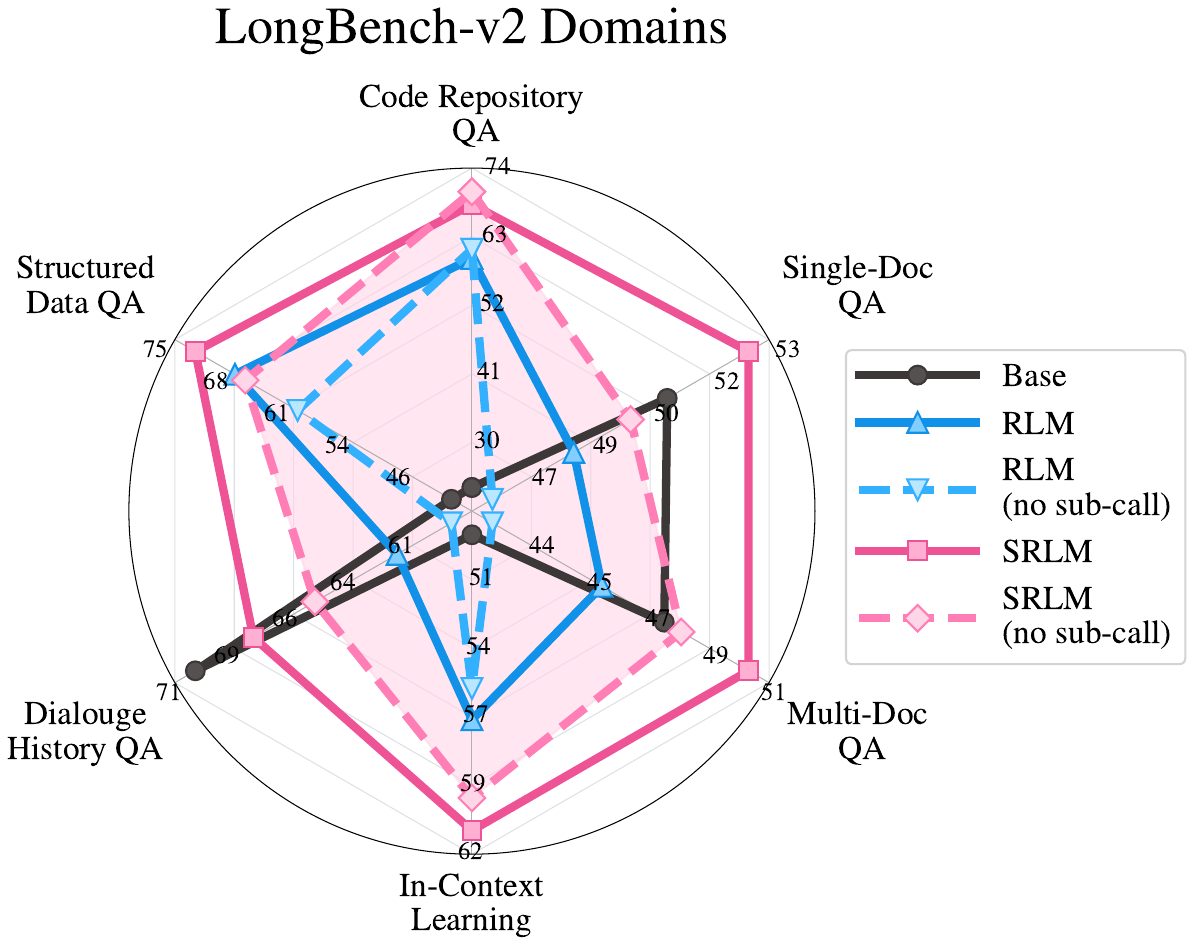}
\caption{
Comparison of \modelname, RLM, and Base LLM across LongBench-v2 domains (averaged across backbone models). In general, \modelname variants show more consistent gains on tasks with different semantic nature. %
}
\label{fig:semantic}
\end{wrapfigure}

Beyond overall long-context performance, it is important to examine how recursion and self-reflection behave across tasks of different natures.
The tasks studied in~\cite{zhang2025recursive} (including OOLONG, BrowseComp+, and LongBench-v2 Code Repository QA) are largely search-oriented. For example, in the LongBench-v2 evaluation, experiments were restricted to the Code QA category, where answering a question typically requires locating specific information across structured repository files. These long contexts are modular and well-organized, making them naturally suitable for recursive and programmatic traversal.
However, LongBench-v2 covers a much broader set of domains beyond code QA, including document QA, dialogue history QA, in-context learning, and others. Many of these tasks are less about finding relevant pieces of information and more about understanding and integrating evidence distributed throughout the entire context. 
To better understand the robustness and consistency of different methods across diverse long-context tasks, we extend our evaluation to these additional LongBench-v2 domains.

Figure~\ref{fig:semantic} shows the performance of each method by domain on LongBench-v2 dataset (averaged across backbones). We observe that the impact of recursion varies substantially with task type. Recursion is particularly more effective for structured, search-oriented tasks such as Code QA and Structured Data QA, but less beneficial for more semantically demanding tasks like Dialogue History QA and Document QA.
In contrast, self-reflection in \modelname variants provides more consistent performance gains across all task categories.
By leveraging the model’s internal uncertainty signals, self-reflective programming offers a stronger semantic steering mechanism for context interaction. This enables it to adapt more effectively to tasks that require different levels and types of semantic understanding.

\subsection{Ablation Study}
\label{sec:exp-ablation}

\begin{figure}[t]
\centering
\includegraphics[width=\linewidth]{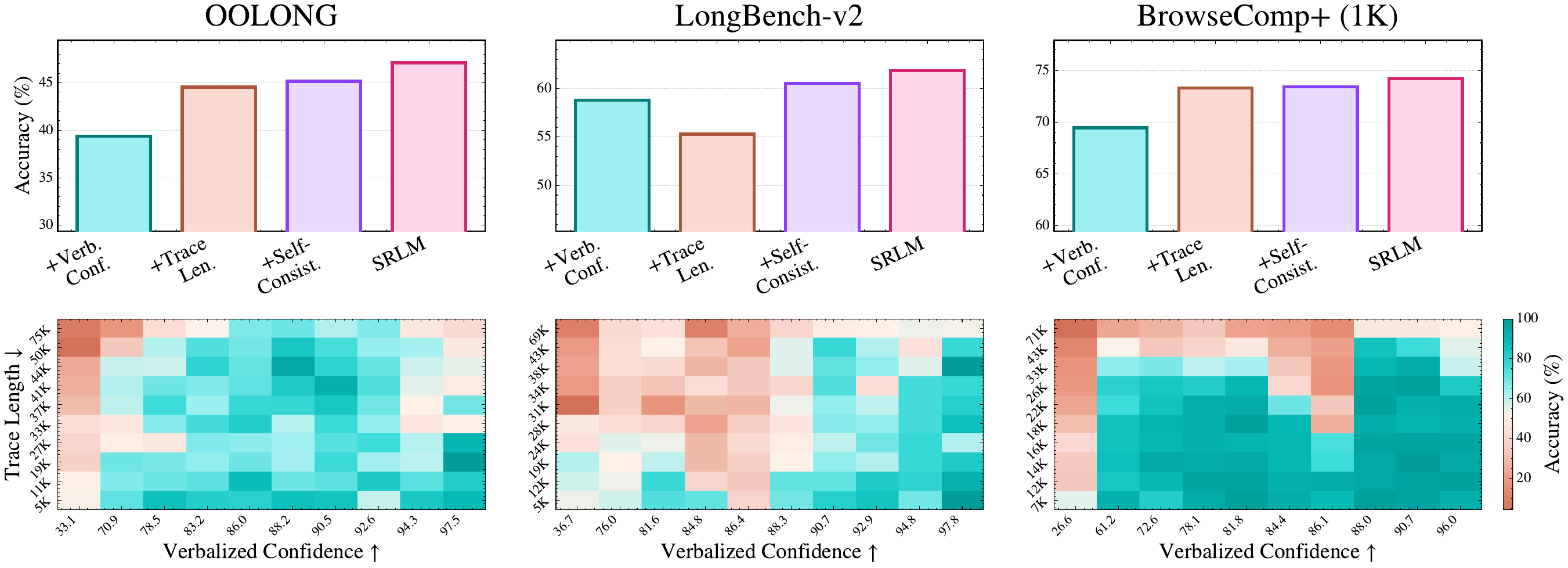}
\caption{
Results of ablation experiments across \modelname's variants (averaged across backbones and recursive/nonrecursive runs). \textbf{Top:} Contribution of each uncertainty signal and their combination in \modelname. \textbf{Bottom:} Complementary effects of semantic and behavioral uncertainty as fine-grained signals guiding self-reflection in \modelname. 
}
\label{fig:ablation}
\end{figure}

In this section, we perform ablations on the self-reflection mechanism of \modelname to understand the contribution of its components. Figure~\ref{fig:ablation} reports results on OOLONG, LongBench-v2, and BrowseComp+ (1K) under long-context settings ($\geq$131K tokens), aggregated across model backbones and runs with and without recursive sub-calls.
\modelname leverages three complementary uncertainty signals during self-reflection: (1)~sampling-based self-consistency, (2)~semantic uncertainty captured through verbalized confidence, and (3)~behavioral uncertainty measured by the length of the reasoning trace.
The top row of Figure~\ref{fig:ablation} shows the impact of each component. As it can be observed, the full \modelname configuration consistently outperforms variants using individual signals, indicating that these uncertainty sources provide complementary benefits.
The bottom row of Figure~\ref{fig:ablation} examines the relation between fine-grained uncertainty metrics (verbalized confidence and reasoning trace length) and accuracy under bins with same samples.
As it can be observed, performance depends jointly on both signals, and their relation with accuracy is not strictly linear. In other words, high confidence or short traces alone do not reliably indicate correctness, however, their combination seem to provide a stronger self-reflection signal in the scope of long-context.

\section{Related Works}

\looseness=-1 \paragraph{Context Window Expansion}

A widely adopted approach for handling long contexts in language models has been to increase the maximum context window through architectural modifications or training strategies. For example, advancements in positional encoding and attention scaling have enabled models to process substantially longer inputs~\cite{chen2023longlora,ding2024longrope,mao2025lift}. 
Beyond these techniques, several architectural directions have been explored to further improve long-context scalability, including sparsity-based mechanisms~\citep{tang2024quest,gao2024seerattention,lai2025flexprefill}, state-space models~\citep{gu2023mamba,dao2024transformers,waleffe2024empirical}, retrieval-augmented fine-tuning~\citep{jin2024long,wang2023augmenting}, and key–value (KV) cache compression techniques~\citep{eyuboglu2025cartridges}. Collectively, these methods aim to improve the scalability of language models with respect to input length and enable effective yet efficient inference over long contexts.
Despite these advances, extending the context window alone does not fully address the challenges of long-context reasoning. Recent empirical studies show that model performance remains constrained by the effective context length. For example, LongBench-v2~\cite{bai2024v2} finds that frontier models with extended context windows still struggle on realistic long-context multitask benchmarks, achieving only modest improvements once input lengths exceed certain context thresholds. These results suggest that scaling the context window, while helpful, is insufficient to guarantee robust reasoning over very large contexts in real-world settings.

\looseness=-1 \paragraph{Agentic Long-Context Approaches.}
There has been an alternative line of research to complement earlier long-context handling approaches with inference-time strategies. Recent works treat long-context as a procedural and agentic problem, leveraging LLMs as tools invoked through programmatic strategies that iteratively interact with context. For example, Recursive Language Model (RLM)~\cite{zhang2025recursive} utilize this idea by treating the context and input prompt as a variable in an external programming environment that can be decomposed, queried, and recursively processed through program execution. RLMs show significant benefits over monolithic prompting and context summarization methods on several long-context benchmarks, demonstrating the effectiveness of programmatic context interaction.
Related approaches include code-execution agents, such as CodeAct~\cite{wang2024executable} which enable iterative code generation and execution for flexible search over context, as well as summarization-based agents such as ReSum~\cite{wu2025resum} that periodically compress interaction histories in agentic web search.
There are also existing works on memory-augmented agents that introduce explicit long-term storage to better extract and retrieve salient information across long interactions, e.g., Mem0 \cite{chhikara2025mem0} and G-Memory \cite{zhang2025g}.
While these agentic and programmatic approaches to long-context tasks show promise, their decisions about what to read, summarize, or revisit typically rely on surface-level heuristics rather than semantic understanding. Consequently, they perform better at structurally localized tasks but struggle with semantically dense ones requiring deep comprehension more than search.

\paragraph{Long-Context Evaluation.} 
A growing body of work in literature has focused on evaluating LLMs under long-context settings. Benchmarks such as LongBench~\cite{bai2024longbenchbilingualmultitaskbenchmark}, and its successor LongBench v2~\cite{bai2024v2} evaluate realistic tasks involving long documents, dialogues, and codebases, revealing substantial performance degradation as context length increases. 
Complementary benchmarks emphasize different stressors. For example, Single Needle in a Haystack (S-NIAH)~\cite{hsieh2024ruler} tasks test retrieval robustness at extreme lengths, where most of the context is irrelevant. In contrast, benchmarks like OOLONG~\cite{bertsch2025oolong} explicitly target long-context reasoning and aggregation, requiring models to process and combine information across all parts of the long iputs. BrowseComp~\cite{wei2025browsecomp} and BrowseComp-Plus~\cite{chen2025browsecomp} also further extend the long-context evaluation to agentic and deep-research settings, where models must persistently navigate, remember, and integrate information across many documents or browsing steps. Collectively, these benchmarks show that long-context reasoning remains a ubiquitous challenge for frontier models across a wide range of diverse and practically important tasks.

\looseness=-1
\paragraph{Confidence Estimation in Large Language Models.}
Estimating model confidence is an important component for enabling reliable reasoning and self-correction in language models. A growing body of work studies how uncertainty signals can be extracted from LLMs without requiring additional fine-tuning. One line of work leverages multiple samples from the model’s predictive distribution to estimate uncertainty via agreement across generations, commonly referred to as sampling-based confidence or self-consistency~\cite{wang2022self, kuhn2023semanticuncertaintylinguisticinvariances, manakul2023selfcheckgptzeroresourceblackboxhallucination}. In this setting, the empirical frequency of an answer across sampled outputs provides a natural estimate of the model’s confidence in that prediction. 
Another direction investigates the model’s ability to explicitly report its own uncertainty. Several recent studies evaluate this called as verbalized confidence, where the model is prompted to provide calibrated confidence scores alongside its predictions~\cite{xiong2023can, yoon2025reasoning}. Empirical comparisons suggest that self-verbalized confidence can serve as a strong and more semantic zero-shot uncertainty estimator and often outperforms approaches based solely on token probabilities or sampling statistics~\cite{tao2025revisitinguncertaintyestimationcalibration}. Other approaches estimate uncertainty using internal model signals such as token likelihoods~\cite{duan2024shifting} or learned probes over hidden representations~\cite{zhang2025reasoning}.
More recently, work in the reasoning literature has identified behavioral signals in the generation process of recent frontier models that correlate with model uncertainty. In particular, several studies observe that incorrect reasoning trajectories tend to be longer and more deliberative than correct ones~\citep{marjanovic2025deepseek, ballon2025relationship}. This phenomenon suggests that reasoning trace length can serve as an implicit proxy for epistemic uncertainty~\citep{devic2025trace, vanhoyweghen2025lexical}. While prior work has primarily exploited this observation to improve reasoning efficiency or reduce unnecessary deliberation~\citep{qu2025optimizing,hassid2025don}, it also provides a useful signal for identifying reliable reasoning trajectories.
Our work builds on these insights and leverages multiple complementary uncertainty signals, including sampling-based consistency, verbalized confidence, and behavioral indicators from reasoning lengths, to guide uncertainty-aware self-reflection during program search for long context interaction.

\section{Conclusion}
In this paper, we study long-context reasoning through the perspective of context-interaction programming. We introduce \modelname, a self-reflective program search framework that uses intrinsic uncertainty signals—self-consistency, reasoning trace length, and verbalized confidence—to guide how models interact with long contexts.
Across diverse benchmarks, context lengths, and backbone models, \modelname consistently improves performance, achieving gains of up to 22\% over the prior state-of-the-art approach RLM.
Our analysis further suggests that recursive decomposition alone is not the main factor behind the performance of Recursive Language Models (RLMs). Instead, the improvements appear to stem from the external programmatic way of handling context interaction.
When guided by self-reflection, these programs provide a more reliable way for models to navigate and reason over long contexts.
In general, \modelname provides more consistent performance gains than RLM across most settings. Notably, it improves performance not only in long-context scenarios but also in shorter contexts within the model's context window, where RLM has been observed to hurt performance. \modelname is also more effective than RLM 
on semantically demanding problems that require deeper contextual comprehension of the context beyond heuristic program search.

Additionally, in this paper, we have employed a relatively simple form of self-reflection based on intrinsic uncertainty signals to guide programmatic context interaction. While effective, this design represents a limitation of our approach and leaves room for future research. Future work could explore richer forms of intrinsic self-reflection within programmatic context-interaction frameworks beyond explicit recursive sub-calls, as well as designs that integrate decision-making with self-reflective signals to enable earlier termination of reasoning and improved control over token usage.
We hope these findings highlight the importance of context-interaction programming for long-context reasoning and suggest that leveraging models’ self-reflective signals is a promising direction for improving long-context capabilities in language models.

\bibliography{refs}
\bibliographystyle{unsrt}

\newpage
\appendix
\onecolumn

\section{Details on Datasets}
\label{app:datasets}

\paragraph{BrowseComp-Plus}\cite{chen2025browsecomp} is a controlled evaluation benchmark for ``deep research'' agents: systems that combine LLM reasoning with search/retrieval tools to answer complex, reasoning-intensive, fact-seeking questions that require combining evidence across multiple sources. Unlike evaluations that rely on live, opaque web search APIs, BrowseComp-Plus fixes the document corpus, enabling reproducible experiments and clearer separation between retrieval failures and reasoning failures.
BrowseComp-Plus is derived from BrowseComp~\cite{wei2025browsecomp}, a benchmark of challenging short-answer browsing problems originally released by OpenAI. BrowseComp was intentionally designed around short, verifiable answers to keep grading straightforward, even though the search process may be difficult.
The BrowseComp-Plus dataset is publicly available\footnote{\url{https://huggingface.co/datasets/Tevatron/browsecomp-plus}} with queries, and the document relevance judgments, including query text and per-query lists of evidence, gold, and negative documents. To reduce benchmark leakage into training corpora of frontier language models and discourage copying benchmark content, the dataset release is obfuscated (with query\_id as the only non-obfuscated field), and includes explicit anti-leakage canary mechanisms. To use the dataset, we need to first decrypt the dataset as suggested \href{https://huggingface.co/datasets/Tevatron/browsecomp-plus}{here}. 

\begin{figure}[t]
\centering
\includegraphics[width=\linewidth]{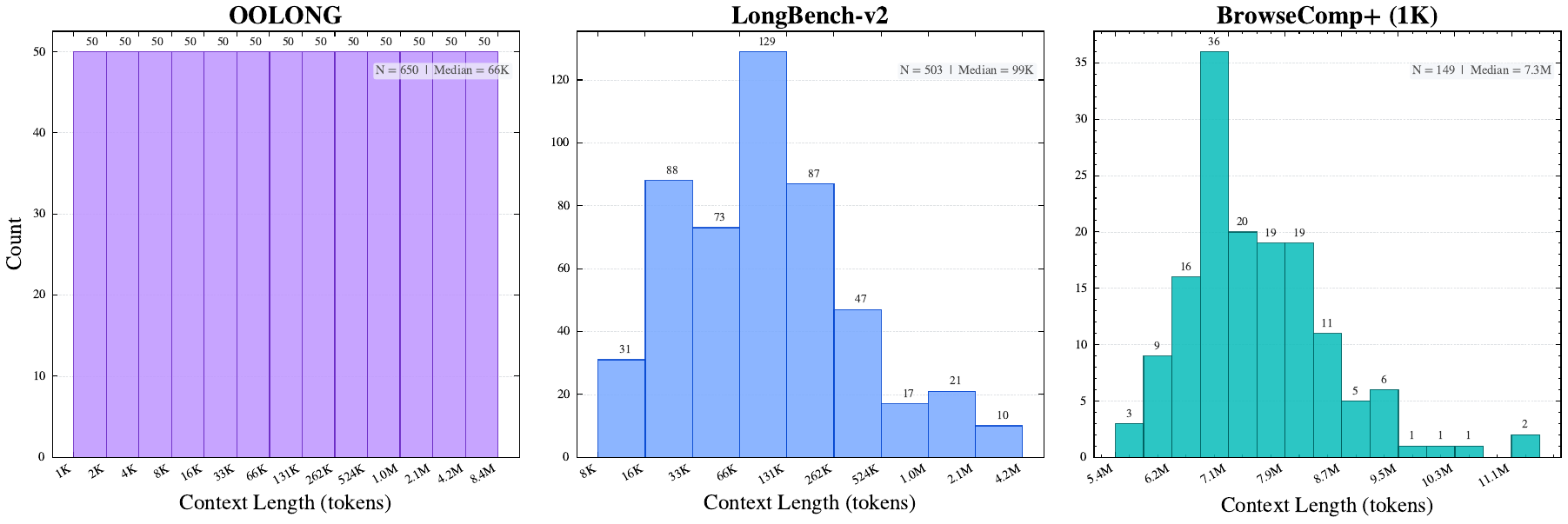}
\caption{
Distribution of input context lengths across different benchmarks: OOLONG, LongBench-v2, and BrowseComp-Plus.
}
\label{fig:app-data}
\end{figure}

In our experiments, we use the ``1K documents'' evaluation setup, following~\cite{zhang2025recursive}: for each question, instead of performing live retrieval, we provide 1,000 randomly selected documents as the long context, with the guarantee that the question’s gold and evidence documents exist within that subset; we then evaluate on 150 randomly sampled questions and report accuracy (percentage of correct final answers). The distribution of context length in terms of tokens across these 150 questions are provided in Figure~\ref{fig:app-data}.
In our setup, we compute robust accuracy using an LLM-as-Judge framework. Instead of relying on exact string matching, the judge compares the model’s generated answer with the provided ground truth using an evaluation prompt (see Section~\ref{app:judge_prompt}), allowing for semantically equivalent or string variations of an answer.

\paragraph{OOLONG}\cite{bertsch2025oolong} is a long-context benchmark designed to evaluate large-scale information aggregation. It is motivated by the observation that many long-context benchmarks reduce to sparse retrieval tasks (i.e., needle-in-a-haystack settings where most of the input is irrelevant noise). In contrast, many real-world long-context tasks require processing a substantial portion of the context and aggregating numerous small decisions to produce a final answer.
OOLONG benchmark is divided into two categories: OOLONG-synth, which consists of ``naturalistic synthetic'' tasks derived from existing in-context learning classification datasets and enables controlled analysis of difficulty factors; and OOLONG-real, which contains downstream aggregation questions over long-form conversational transcripts, designed to be less easily decomposed into independent in-context learning examples.

\begin{figure}[t]
\centering
\includegraphics[width=\linewidth]{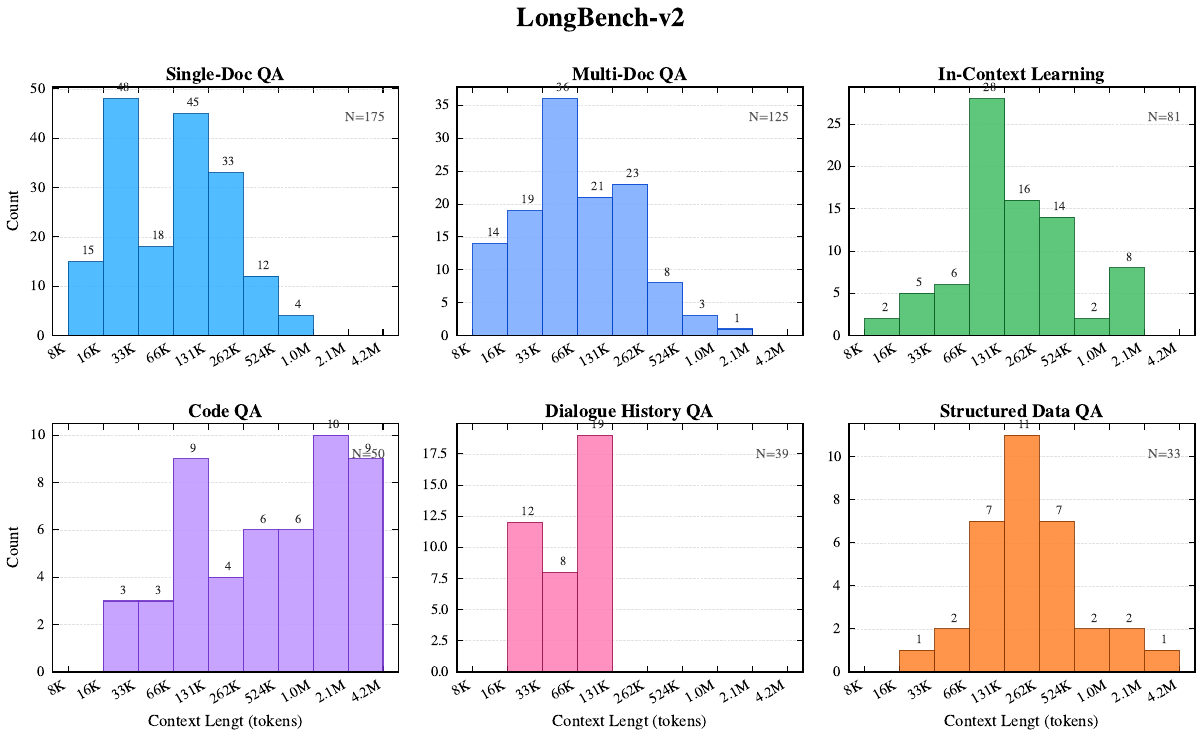}
\caption{
Context length distributions for different task categories in LongBench-v2 benchmark dataset.
}
\label{fig:app-data-domains}
\end{figure}

In our experiments, we focus on the OOLONG-synth \texttt{trec\_coarse} configuration, following~\cite{zhang2025recursive}. We evaluate 650 tasks across context lengths ranging from 1K to 8M tokens (50 tasks per context length as shown in Figure~\ref{fig:app-data}) and report accuracy following the benchmark's original scoring protocol, including exponential partial credit for numeric outputs.
For categorical outputs (e.g., labels, dates, user IDs, or comparisons), correctness is determined by exact match with the ground truth or by judge-based equivalence assessment (with same prompt from Section~\ref{app:judge_prompt}). For numeric outputs, the score is computed as $0.75^{|y-\hat{y}|}$ which assigns partial credit as predictions approach the true value.
We use the publicly available dataset \footnote{\url{https://huggingface.co/datasets/oolongbench/oolong-synth}}
together with the official open-source code\footnote{\url{https://github.com/abertsch72/oolong}} for experiments.

\paragraph{LongBench-v2}\cite{bai2024v2} is an extended version of the LongBench benchmark~\cite{bai2024longbenchbilingualmultitaskbenchmark}, designed to evaluate the ability of large language models to perform reasoning and comprehension over diverse realistic long contexts. 
To make evaluation more robust and less sensitive to generation formatting, the benchmark standardizes tasks into multiple-choice questions (A/B/C/D), allowing consistent comparison across models with different behaviors.
LongBench-v2 contains 503 evaluation instances with context lengths ranging from 8K to 4M tokens. The overall context length distribution of this dataset is shown in Figure~\ref{fig:app-data}. The benchmark spans a wide spectrum of realistic long-context scenarios and is organized into six task categories: single-document question answering, multi-document question answering, long in-context learning, dialogue history understanding, code repository understanding, and long structured data understanding. These categories collectively capture a diverse set of long-context reasoning patterns, including extracting from and comprehending lengthy documents, synthesizing and finding evidence across multiple sources, reasoning over extended conversational histories, understanding large software repositories, and querying or aggregating information from large structured tables. The detailed context length distributions for each of these domain categories are provided in Figure~\ref{fig:app-data-domains}.

To assess how different long-context methods generalize across diverse domains, we evaluate on all task categories of the LongBench-v2 dataset in our experiments. This differs from prior work~\cite{zhang2025recursive}, which focuses only on the CodeQA subset of the benchmark. Using the full benchmark enables a more comprehensive evaluation of long-context capability across heterogeneous domains and task semantics. In our experiments, we use the official LongBench-v2 dataset which is publicly available\footnote{\url{https://huggingface.co/datasets/zai-org/LongBench-v2}} together with the benchmark's open-source code for evaluation\footnote{\url{https://github.com/THUDM/LongBench}}. Simialr to other datasets, to evaluate correctness of answers in this benchmark, we also use the judge with prompt as in Section~\ref{app:judge_prompt}.

\section{Prompt Details}

\subsection{Verbalized Confidence Elicitation}
\label{app:confidence_prompt}
At each intermediate generation step $t$ of every candidate program $p^{(k)}$, we elicit the model's self-assessed confidence by appending a fixed structured instruction to the generation prompt. The instruction is designed to (i)~require the model to produce a parseable confidence score, (ii)~enforce a consistent format across all steps and all programs, and (iii)~encourage nuanced and calibrated self-assessment rather than coarse or overconfident reporting.
The following instruction is appended verbatim to the model's prompt for every generation. 
For the rare cases that the verbalized confidence is not provided by model, the average of verbalized confidence of other steps in the same trajectory is used to fill the gap.

\begin{tcolorbox}[breakable, colback=customwhite, colframe=customgray,title=Confidence Prompt, fontupper=\fontfamily{pcr}\selectfont\scriptsize]
\begin{verbatim}
IMPORTANT: At the end of your response, you MUST include 
your confidence in your answer using this exact JSON format 
on a new line:
```json
{"confidence": <number between 0.000 and 100.000 
                with up to 3 decimal points>}
```
Be precise and nuanced in your confidence assessment.
\end{verbatim}
\end{tcolorbox}

\subsection{Evaluation Judge}
\label{app:judge_prompt}

To compute final task accuracy across all datasets used in our experiments, we employ an LLM-as-Judge evaluation procedure. Many long-context benchmarks contain answers that may appear in slightly different textual forms while still being semantically correct (e.g., variations in phrasing, formatting differences, or additional explanatory details). Relying solely on exact string matching can therefore incorrectly penalize some of the valid and semantically equivalent answers. To address this issue, we use an automated judge model with the prompt below that compares the model-generated responses with the ground-truth answer and determines correctness based on semantic equivalence rather than strict lexical matching.

\begin{tcolorbox}[breakable, colback=customwhite, colframe=customgray,
title=Grading Prompt, fontupper=\fontfamily{pcr}\selectfont\scriptsize]
\begin{verbatim}
Judge whether the following [response] to [question] is correct or not
based on the precise and unambiguous [correct_answer] below.

[question]: {QUESTION}

[response]: {RESPONSE}

[correct_answer]: {CORRECT_ANSWER}

Your judgement must be in the format and criteria specified below:

extracted_final_answer: The final exact answer extracted from the [response].

correct_answer: Repeat the [correct_answer] given above.

reasoning: Explain why the extracted_final_answer is correct or incorrect based on
[correct_answer], in the context of this [question]. You should judge
whether the extracted_final_answer is semantically equivalent to
[correct_answer], allowing the extracted_final_answer to be string
variations of [correct_answer]. You should also allow the
extracted_final_answer to be more precise or verbose than
[correct_answer], as long as its additional details are correct.
Do not comment on any background to the problem, do not attempt to
solve the problem, do not argue for any answer different than
[correct_answer], focus only on whether the answers are
semantically equivalent.

correct: Answer 'yes' if extracted_final_answer matches the [correct_answer]
given above, or is within a small margin of error for numerical
problems. Answer 'no' otherwise, i.e. if there is any inconsistency,
ambiguity, non-equivalency, or if the extracted answer is incorrect.

confidence: The extracted confidence score between 0|%

\end{verbatim}
\end{tcolorbox}

\section{Additional Results}

\subsection{Detailed Results Across Context Lengths}
\label{app:context_details}

\begin{figure}[t]
\centering
\includegraphics[width=\linewidth]{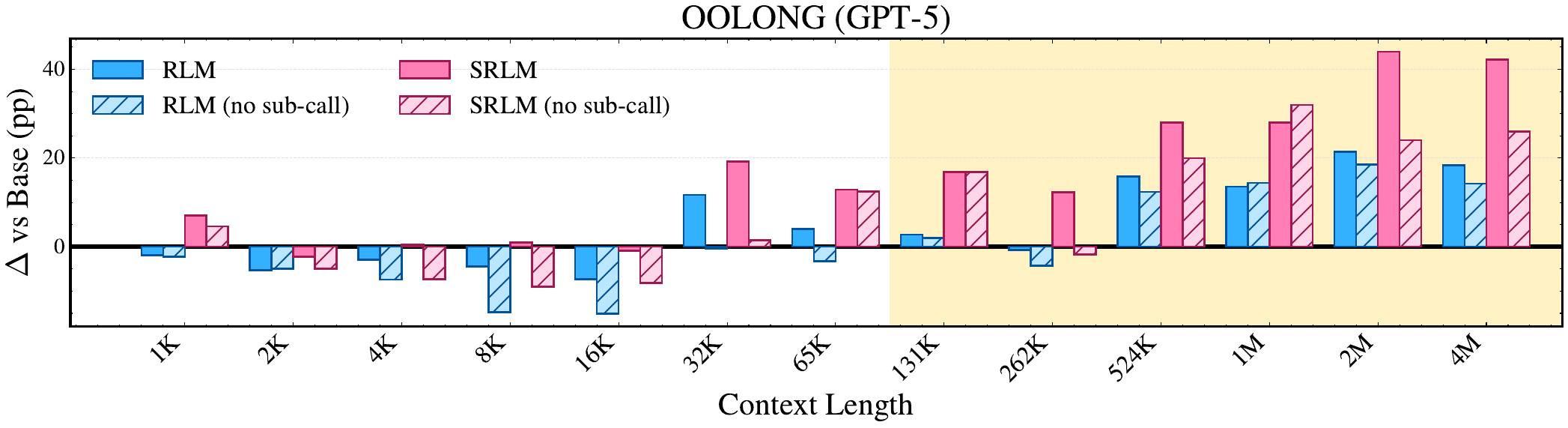}

\vspace{0.5em}

\includegraphics[width=\linewidth]{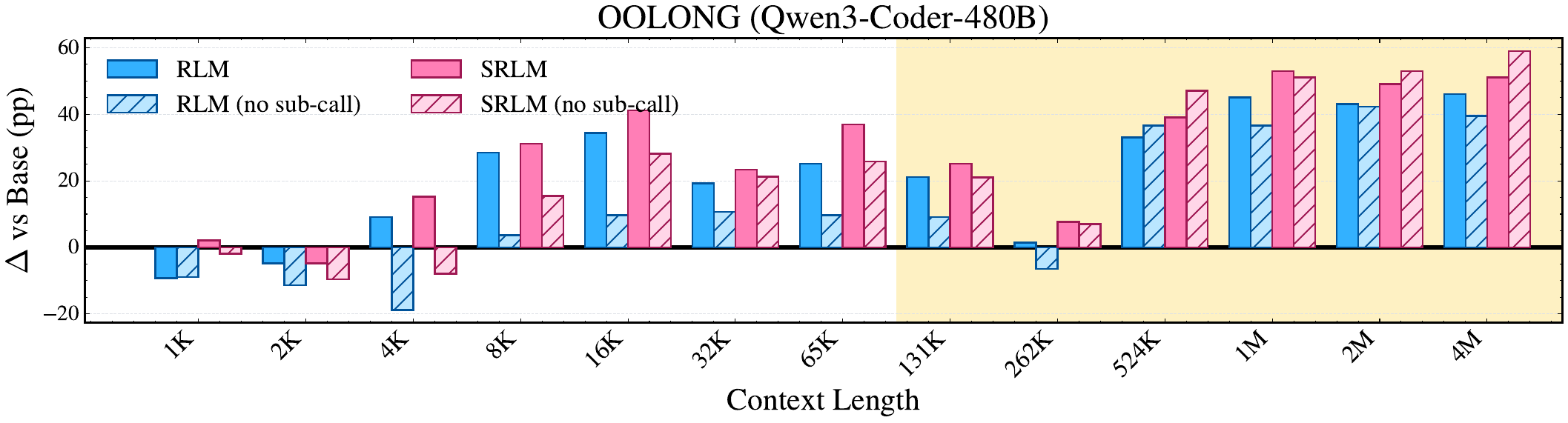}
\caption{
Performance gains over the base model across context lengths on the OOLONG benchmark. Bars show accuracy change vs. base for each context-length bin. Results are shown for GPT-5 (\textbf{top}) and Qwen3-Coder-480B (\textbf{bottom}). The shaded yellow region marks contexts near or beyond the context window limits ($\geq 131$K tokens).
}
\label{fig:oolong-gap}
\end{figure}

\begin{figure}[t]
\centering
\includegraphics[width=\linewidth]{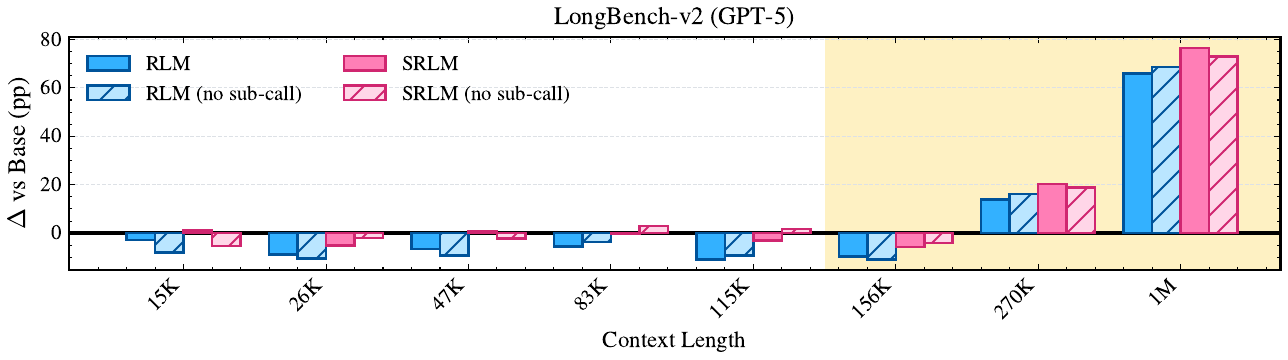}

\vspace{0.5em}

\includegraphics[width=\linewidth]{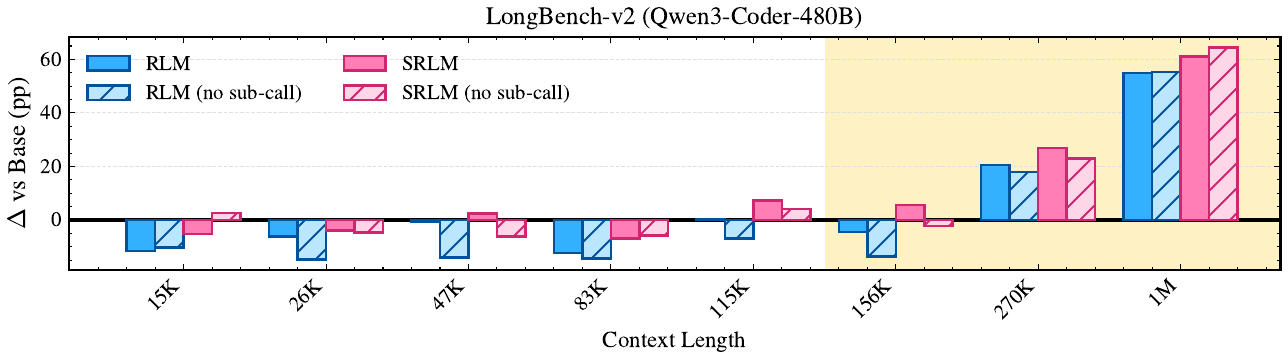}
\caption{
Performance gains over the base model across context lengths on the LongBench-v2 benchmark. Bars show accuracy change vs. base for each context-length bin with balanced samples. Results are shown for GPT-5 (\textbf{top}) and Qwen3-Coder-480B (\textbf{bottom}) backbones. The shaded yellow region marks contexts near or beyond the context window limits ($\geq 131$K tokens).
}
\label{fig:longbench-gap}
\end{figure}

\begin{figure}[t]
\centering
\includegraphics[width=\linewidth]{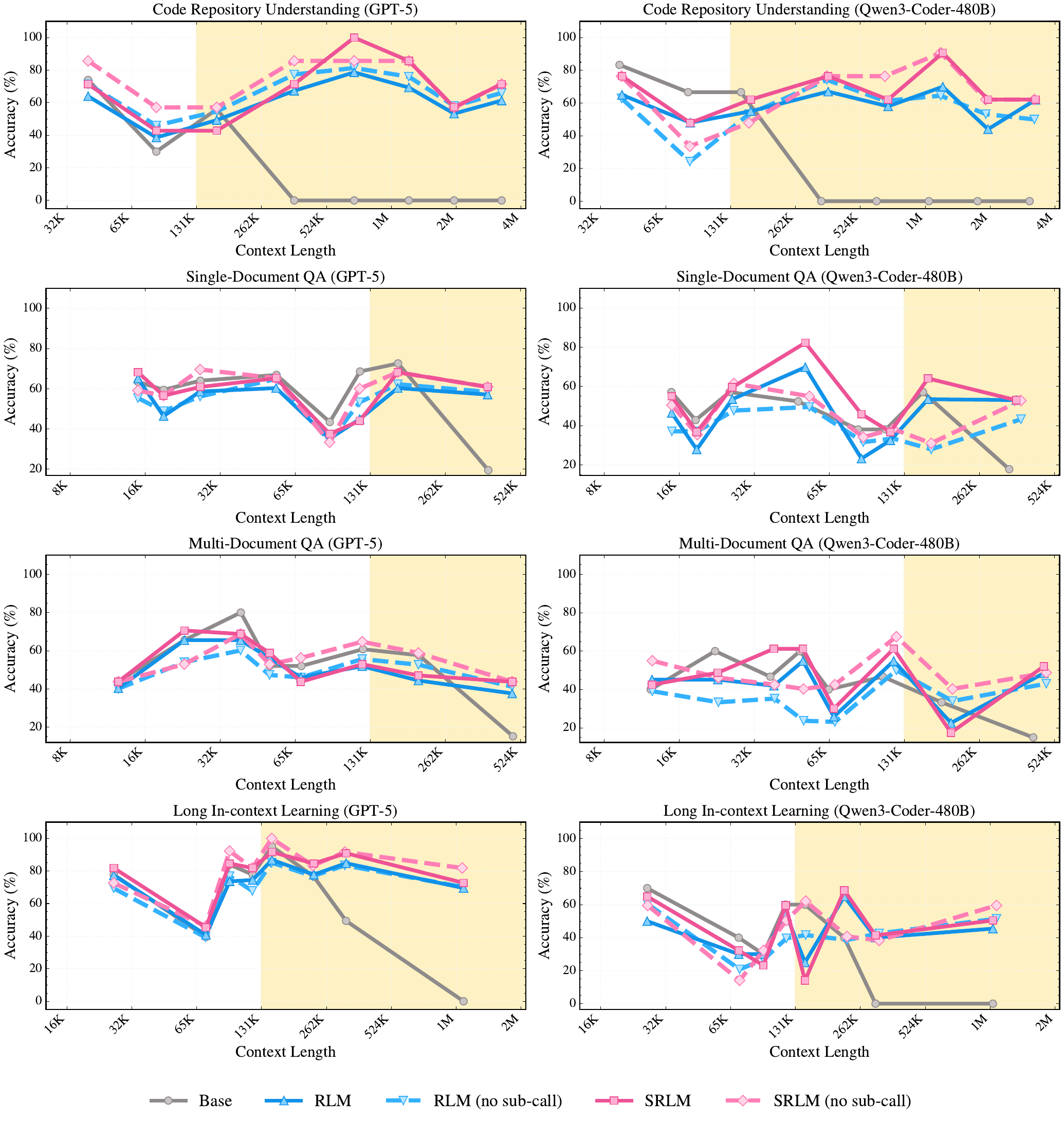}
\caption{
Accuracy versus context length across LongBench-v2 task categories. Results for GPT-5 (\textbf{left}) and Qwen3-Coder-480B (\textbf{right}) backbones, with each subplot showing performance across context scales for a specific task domain.
}
\label{fig:longbench-domains-acc}
\end{figure}

\begin{figure}[h]
\centering
\includegraphics[width=\linewidth]{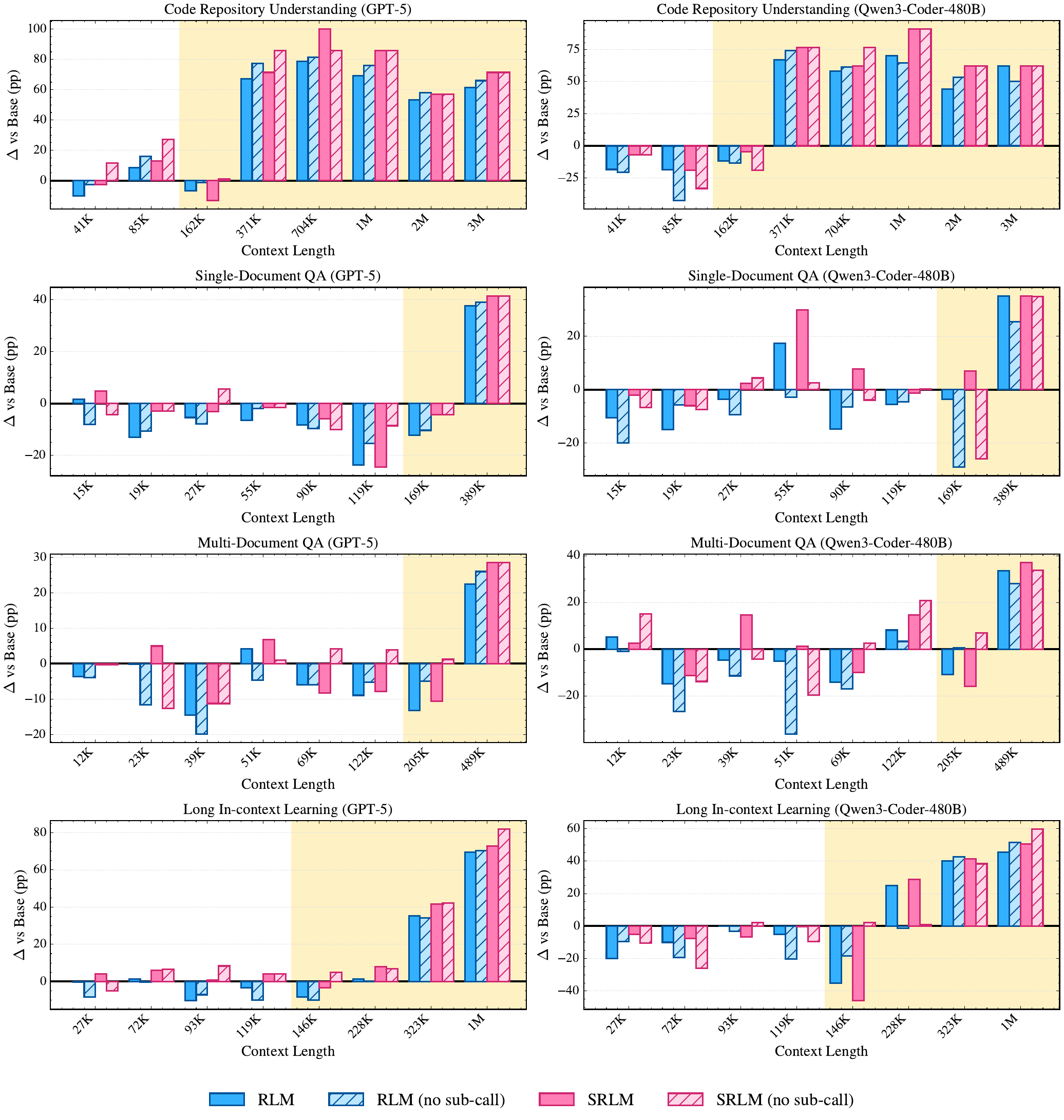}
\caption{
Improvement over the base model across context lengths for LongBench-v2 task domains. Results for GPT-5 (\textbf{left}) and Qwen3-Coder-480B (\textbf{right}), with each subplot showing performance change across context scales for a specific task domain.
}
\label{fig:longbench-domains-gap}
\end{figure}

\paragraph{Improvement over the base model.}
In this section, we provide a more detailed view of the context-length experiments summarized in Section~\ref{sec:exp-context}. While the main paper presents aggregated results for comparison to base across context splits and task domains in LongBench-v2 dataset, Figure~\ref{fig:oolong-gap}-\ref{fig:longbench-domains-gap} show performance trends at a finer granularity across context-length bins and different task categories. These additional results help illustrate how different approaches behave as the input context grows from moderately long inputs to ver long multi-million token sequences.
Figures~\ref{fig:oolong-gap} and~\ref{fig:longbench-gap} present the detailed improvement relative to the base model ($\Delta$ vs.\ base) for each context-length bin on the OOLONG and LongBench-v2 benchmarks, respectively. Each bin contains approximately the same number of evaluation instances to ensure fair comparison across context scales. The plots report the performance difference between each method and the base model, allowing us to directly observe whether a method improves or degrades performance at different context lengths. As it can be shown, the context length is separated into contexts within (<131K) and near/beyond
($≥131$K) the model's context window with the highlighted yellow background.

Consistent with the trends observed in Figure~\ref{fig:context} in the main paper, the advantage of \modelname\ becomes more pronounced at longer contexts. In both benchmarks and across both backbone models, the improvement over the base model increases as context length grows. In contrast, RLM mostly underperforms the base model at shorter context lengths, indicating that recursive decomposition may introduce unnecessary overhead when the context already fits within the model's window.

\paragraph{Task-domain analysis.}
To further understand how long-context methods behave across different types of reasoning tasks, Figures~\ref{fig:longbench-domains-acc} and~\ref{fig:longbench-domains-gap} provide detailed breakdowns by task category for the context analysis performance on LongBench-v2 benchmark. These plots report both the absolute accuracy and the improvement relative to the base model across context lengths detailed for each task domain.
The results reveal that the benefits self-reflection in \modelname\ are consistent across diverse tasks, including code repository understanding, single-document question answering, multi-document question answering, and long in-context learning. In particular, improvements become more pronounces once the context length approaches or exceeds the context window limits of the underlying model (shaded yellow region).

\subsection{Detailed Ablation Results}
\label{app:ablation_details}

This section provides a detailed breakdown of the ablation experiments summarized in Section~\ref{sec:exp-ablation}. While the main paper reports aggregated results across recursive/nonrecursive runs and backbones, the figures in this appendix present the results separately for each backbone model, dataset, and execution setting. These analyses allow us to better understand how individual uncertainty signals contribute to the performance of \modelname\ under different conditions.

\begin{figure}[t]
\centering
\begin{subfigure}{\linewidth}
\centering
\caption{Recursive}
\includegraphics[width=\linewidth]{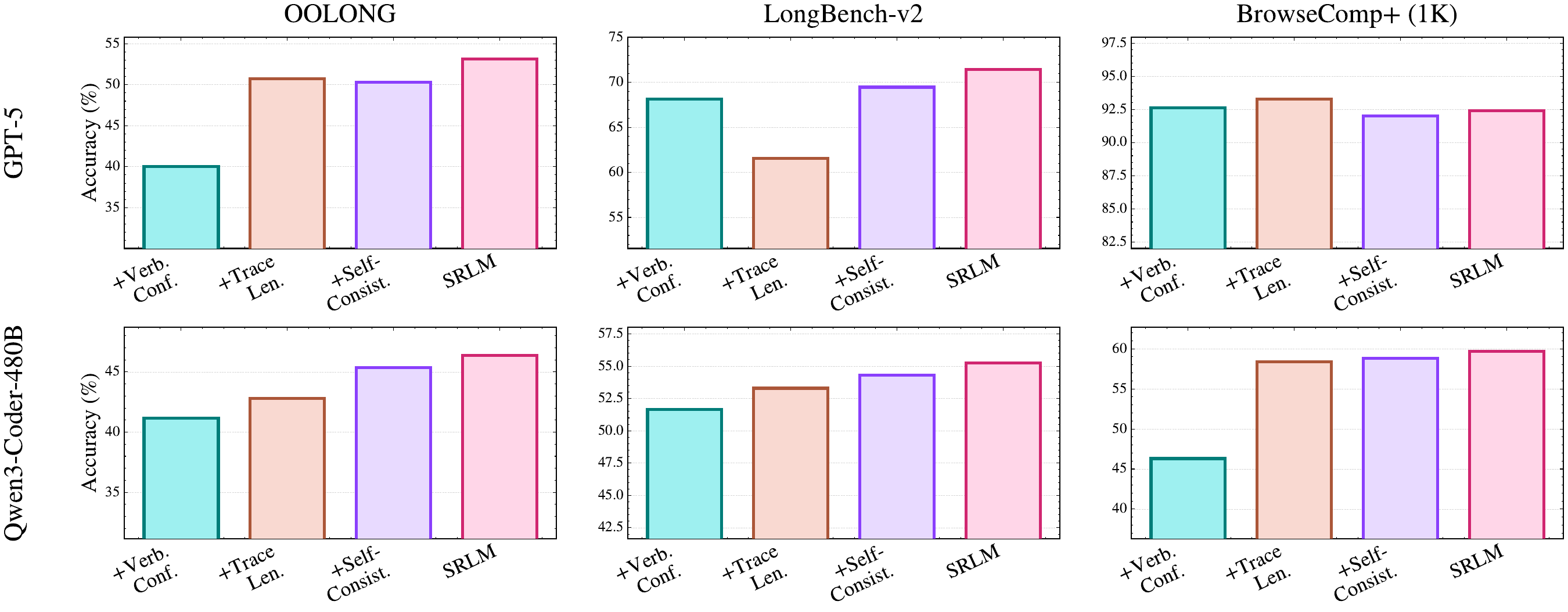}
\end{subfigure}

\vspace{0.5em}

\begin{subfigure}{\linewidth}
\centering
\caption{No Sub-call}
\includegraphics[width=\linewidth]{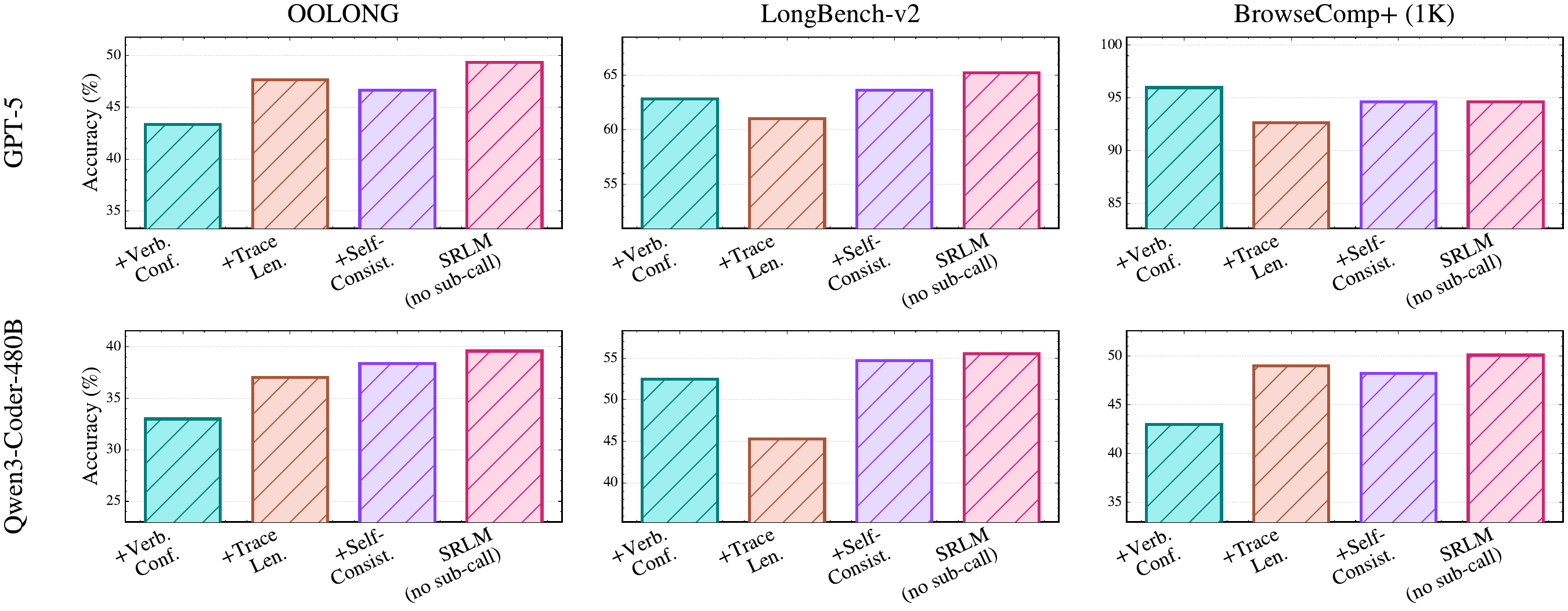}
\end{subfigure}

\caption{
Ablation of uncertainty signals in \modelname’s self-reflection mechanism. Results for GPT-5 and Qwen3-Coder-480B backbones on OOLONG, LongBench-v2, and BrowseComp+ (1K) benchmarks with context $\geq131$K. \textbf{(a) With} recursive sub-calls; \textbf{(b) Without} recursive sub-calls. 
}
\label{fig:app-ablation1}

\end{figure}

Figure~\ref{fig:app-ablation1} shows the detailed analysis for contribution of each uncertainty signal used in the self-reflection mechanism of \modelname. The results are reported separately for GPT-5 and Qwen3-Coder-480B backbones, as well as for runs with recursive sub-calls and runs without sub-calls.
Across all datasets and model backbones, the full \modelname\ configuration consistently achieves the best performance. Variants that rely on individual uncertainty signals alone show smaller improvements compared to the combination in \modelname for all the queries with context length $\geq131$K. This observation confirms that the three uncertainty signals capture complementary aspects of model uncertainty during reasoning and this is true across all backbones and runs. 
The detailed plots also reveal that the relative contribution of each signal can vary depending on the backbone model and dataset. For example, reasoning trace length tends to provide stronger signals in some datasets, whereas verbalized confidence tends to be more informative on others when the model's internal calibration is reliable. However, across all settings, combining the signals through the self-reflection mechanism consistently provides the best performance.

\begin{figure}[t]
\centering
\begin{subfigure}{\linewidth}
\centering
\caption{Recursive}
\includegraphics[width=\linewidth]{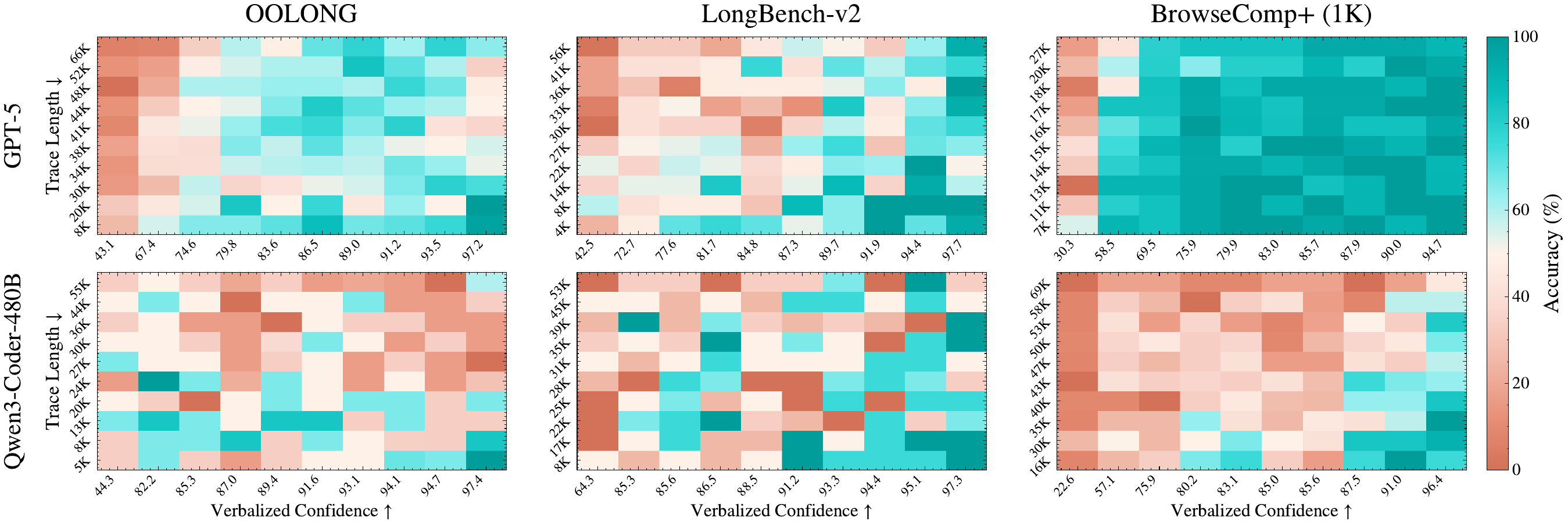}
\end{subfigure}

\vspace{0.5em}

\begin{subfigure}{\linewidth}
\centering
\caption{No Sub-call}
\includegraphics[width=\linewidth]{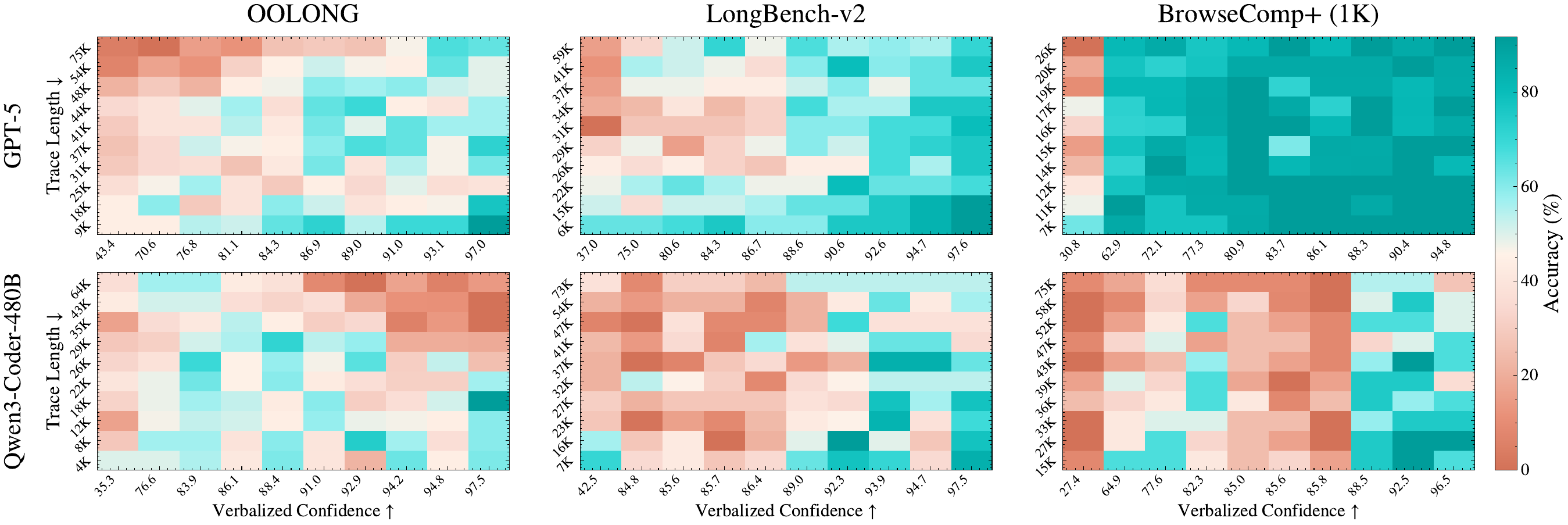}
\end{subfigure}

\caption{
Analysis of the relationship between verbalized confidence, reasoning length, and accuracy. Heatmaps show accuracy across equal-sample bins of confidence (x-axis) and reasoning length (y-axis). Results are reported for GPT-5 and Qwen3-Coder-480B backbones on the OOLONG, LongBench-v2, and BrowseComp+ (1K) benchmarks with context $\geq131$K. \textbf{(a) With} recursive sub-calls; \textbf{(b) Without} recursive sub-calls.
}
\label{fig:app-ablation2}
\end{figure}

Figure~\ref{fig:app-ablation2} further analyzes the interaction between fine-grained signals of verbalized confidence and reasoning trace length. For this analysis, predictions are grouped into bins with approximately equal numbers of samples based on their confidence scores and reasoning trace lengths. Each heatmap cell reports the empirical accuracy obtained for a given combination of these two signals.
The heatmaps reveal several interesting patterns. For example, the relation between verbalized confidence and accuracy is not strictly monotonic, i.e., predictions with high confidence do not always correspond to correct answers. Similarly, shorter or longer reasoning traces alone do not reliably indicate correctness. Instead, the highest accuracy regions often appear when both signals align and their joint combination is considered. 
By jointly considering semantic uncertainty (verbalized confidence) and behavioral uncertainty (reasoning trace characteristics), \modelname\ is able to provide more reliable self-reflection for context-interaction programs in the scope of long-context.

\end{document}